\documentclass[journal]{IEEEtran}
\usepackage{amsmath,amsfonts}
\usepackage{bm}
\usepackage{algorithmic}
\usepackage{algorithm}
\usepackage{array}
\usepackage[caption=false,font=normalsize,labelfont=sf,textfont=sf]{subfig}
\usepackage[export]{adjustbox}
\usepackage{textcomp}
\usepackage{stfloats}
\usepackage{url}
\usepackage{hyperref}
\usepackage{verbatim}
\usepackage{graphicx}
\usepackage{cite}
\usepackage{amssymb}
\usepackage{booktabs, multirow}
\usepackage[normalem]{ulem}
\usepackage{multirow}
\usepackage{diagbox}
\usepackage{makecell}
\usepackage{pifont}
\usepackage[table,dvipsnames]{xcolor}
\usepackage{pgf}
\usepackage{collcell}  

\newcommand{\cmark}{\cellcolor{green!5}\textcolor{ForestGreen}{\normalsize\textbf{\ding{51}}}}%
\newcommand{\bcmark}{\cellcolor{blue!5}\textcolor{NavyBlue}{\normalsize\textbf{\ding{51}}}}%
\newcommand{\ccmark}{\cellcolor{Purple!5}\textcolor{Purple}{\normalsize\textbf{\ding{51}}}}%
\newcommand{\xmark}{\cellcolor{red!5}\textcolor{BrickRed}{\normalsize\textbf{\ding{55}}}}%
\newcommand{\gcmark}{\cellcolor{gray!5}\textcolor{gray}{\normalsize\textbf{\ding{51}}}}%
\newcommand{\mcmark}{\cellcolor{Melon!5}\textcolor{Melon}{\normalsize\textbf{\ding{51}}}}%

\newcommand{\cellcolorval}[1]{%
  \pgfmathsetmacro{\val}{1-#1}
  \ifdim\val pt<.5pt
    \pgfmathsetmacro{\g}{2*\val}%
    \pgfmathsetmacro{\r}{1}%
    \pgfmathsetmacro{\b}{0}%
  \else
    \pgfmathsetmacro{\t}{2*(\val-0.5)}%
    \pgfmathsetmacro{\r}{1-\t}%
    \pgfmathsetmacro{\g}{1}%
    \pgfmathsetmacro{\b}{0}%
  \fi
  \pgfmathsetmacro{\lum}{0.299*\r+0.587*\g+0.114*\b}%
  \ifdim\lum pt<0.5pt \def\textcol{white}\else\def\textcol{black}\fi
  \edef\colorcmd{\noexpand\cellcolor[rgb]{\r,\g,\b}}%
  \pgfmathsetmacro{\val}{#1}
  \colorcmd\textcolor{\textcol}{\pgfmathprintnumber[fixed,precision=2]{\val}}%
}

\newcommand{\mytextcircled}[1]{%
  \raisebox{.5pt}{\textcircled{\raisebox{-0.9pt}{#1}}}%
}

\begin{document}

\title{Apple's Synthetic Defocus Noise Pattern: Characterization and Forensic Applications}

\author{David Vázquez-Padín,~\IEEEmembership{Member,~IEEE,} Fernando Pérez-González,~\IEEEmembership{Fellow,~IEEE,} and Pablo Pérez-Miguélez%
\thanks{Work partially funded by the EU (Next Generation) through the ``Plan de Recuperación, Transformación y Resiliencia'' under project FELDSPAR: ``Federated Learning with Model Ownership Protection and Privacy Armoring'' (Grant MCIN/AEI/10.13039/501100011033) and by Xunta de Galicia and the European Regional Development Fund, under projects ED431C 2021/47 and ED431C 2025/41. Funding for open access charge: Universidade de Vigo/CISUG.}
}



\maketitle

\begin{abstract}
iPhone portrait-mode images contain a distinctive pattern in out-of-focus regions simulating the bokeh effect, which we term Apple's \emph{Synthetic Defocus Noise Pattern} (SDNP). If overlooked, this pattern can interfere with blind forensic analyses, especially PRNU-based camera source verification, as noted in earlier works. Since Apple's SDNP remains underexplored, we provide a detailed characterization, proposing a method for its precise estimation, modeling its dependence on scene brightness, ISO settings, and other factors. Leveraging this characterization, we explore forensic applications of the SDNP, including traceability of portrait-mode images across iPhone models and iOS versions in open-set scenarios, assessing its robustness under post-processing. Furthermore, we show that masking SDNP-affected regions in PRNU-based camera source verification significantly reduces false positives, overcoming a critical limitation in camera attribution, and improving state-of-the-art techniques.
\end{abstract}

\begin{IEEEkeywords}
Image forensics, source camera verification, PRNU, portrait mode, depth, computational imaging, Apple.
\end{IEEEkeywords}

\section{Introduction}
\label{sec:intro}

\IEEEPARstart{T}{he} widespread adoption of computational photography in modern smartphones, especially with the integration of Artificial Intelligence (AI) algorithms, poses considerable challenges for classic methods in multimedia forensics, particularly in the area of image forensics. Examples of computational imaging techniques include image stitching to create panoramic photos, High Dynamic Range (HDR) imaging to extend the camera's native luminosity range in a single shot, ``night mode'' to enhance low-light imaging, and ``portrait mode'' for simulating a shallow depth of field and mimic a realistic bokeh. Digital images produced through any of these computational imaging modes undergo complex processing chains, some involving black-box systems. These chains deviate significantly from well-established image acquisition models that enable source camera verification (also known as attribution) using Photo Response Non-Uniformity (PRNU), detect image manipulations by analyzing demosaicing patterns and/or JPEG compression traces, and more.

Although computational imaging modes are not inherently malicious, Iuliani \emph{et al.} \cite{IULIANI_2021} showed that Apple's portrait mode can mislead forensic analysis, causing false positives in camera source attribution. Baracchi \emph{et al.} \cite{BARACCHI_2021} previously found traditional PRNU-based methods \cite{CHEN_2007} ineffective for iPhone X portrait images and proposed using depth maps to remove the so-called Non-Unique Artifacts (NUAs). Similar issues were later observed in Samsung and Huawei devices in \cite{ALBISANI_2021}, with partial mitigation via a SPAM classifier \cite{BHAT_2022}. More recently, misattributions in other devices were further studied in \cite{MONTIBELLER_2024}.

Both \cite{ALBISANI_2021} and \cite{BHAT_2022} recommend isolating images generated with computational imaging modes, which can be done using McCloskey's method \cite{MCCLOSKEY_2022} to detect focus manipulations like portrait mode. However, forensic practitioners may still face cases where only portrait images are available. As explored in this paper, forensic identification and characterization of computational imaging patterns can help not only recognize processed images, but also isolate affected regions within an image to mitigate errors in camera source attribution and detect inconsistencies where known patterns have been embedded.

Inspired by the call to action in \cite{IULIANI_2021}, we have chosen to thoroughly investigate Apple's portrait mode. This mirrors the approach taken by Butora and Bas in \cite{BUTORA_2024}, who examined the pattern introduced by Adobe in the development of raw or 16-bit images, which resulted in PRNU collisions. By modeling the so-called Adobe pattern, they significantly reduced collisions and later proposed a method to locally detect the presence of this pattern (at a resolution of $128 \times 128$ pixels) across an entire image in \cite{BUTORA_2024b}. Our focus on Apple's portrait mode is motivated by Apple's position as a leading smartphone manufacturer, consistently ranking among the top sellers since 2022 \cite{COUNTERPOINT_2024} and leading smartphone sales with the iPhone 15 (released in September 2023) since Q4 2023 \cite{COUNTERPOINT_MODEL_SALES_2024}.

Our work provides an in-depth characterization of the pattern embedded by Apple in portrait-mode images, first exposed in \cite{BARACCHI_2021}, which we term the Synthetic Defocus Noise Pattern (SDNP), offering tools and insights to effectively handle portrait images in forensic applications. Specifically, our contributions include:
\begin{itemize}
        \item Methods for SDNP extraction using two lighting modes.
        \item Characterization of SDNP's dependence on scene luminance and ISO settings.
        \item Tracking of SDNP variations across different image resolutions, iPhone models, and iOS versions.
        \item Leveraging extracted SDNPs to reduce the PRNU collisions noted in \cite{IULIANI_2021} and improve state-of-the-art tools \cite{BARACCHI_2021}.
        \item Assessing SDNP detection robustness under complex post-processing conditions, such as image sharing via WhatsApp.
\end{itemize}

The paper is organized as follows: Sect.~\ref{subsec:notation} introduces the notation, followed by a review of PRNU-based source verification and Apple's portrait mode in Sect.~\ref{sec:preliminaries}. SDNP extraction and characterization are covered in Sects.~\ref{sec:BP_extraction} and~\ref{sec:estimation_BP_scaling_factors}, with pattern variations discussed in Sect.~\ref{sec:BP_variations}. Sect.~\ref{sec:BP_applications} presents forensic applications, and experimental results are reported in Sect.~\ref{sec:exp_results}. Conclusions are drawn in Sect.~\ref{sec:conclusions}.

\subsection{Notation}
\label{subsec:notation}

Matrices are denoted by bold uppercase letters, vectors by bold lowercase letters, and scalars by regular (non-bold) letters. Unless otherwise stated, matrices are real-valued with dimensions $H\times W$. The $(i,j)$th element of a matrix $\mathbf{A}$ is written as $A_{i,j}$, where $0\leq i\leq H-1$ and $0\leq j\leq W-1$. The total number of elements of $\mathbf{A}$ is $N\triangleq H\cdot W$. The Frobenius inner product of matrices $\mathbf{A}$ and $\mathbf{B}$ is $\langle\mathbf{A},\mathbf{B}\rangle_{\text{F}}\triangleq\sum_{i=0}^{H-1}\sum_{j=0}^{W-1}A_{i,j}B_{i,j}$, and the Frobenius norm of $\mathbf{A}$ is $\|\mathbf{A}\|_{\text{F}}\triangleq\sqrt{\langle\mathbf{A},\mathbf{A}\rangle_{\text{F}}}$. The Hadamard product (i.e., element-wise product) of $\mathbf{A}$ and $\mathbf{B}$, denoted $\mathbf{A}\circ\mathbf{B}$, results in a matrix of the same dimension as the operands, with elements $\left(\mathbf{A}\circ\mathbf{B}\right)_{i,j} = A_{i,j}\cdot B_{i,j}$. The Hadamard inverse of $\mathbf{A}$, denoted $\mathbf{A}^{\circ-1}$, is defined element-wise as $\left(\mathbf{A}^{\circ-1}\right)_{i,j}=A_{i,j}^{-1}$. Scalar-valued functions are denoted by lowercase letters, e.g., $f(\cdot)$, while uppercase letters, e.g., $F(\cdot)$, are used for matrix-valued functions. The sample mean of $\mathbf{A}$ is defined as $\mu(\mathbf{A})\triangleq \langle\mathbf{A},\mathbf{1}\rangle_{\text{F}}/N$, where $\mathbf{1}$ is an $H\times W$ matrix of ones. The sample standard deviation of $\mathbf{A}$ is defined as $\sigma\left(\mathbf{A}\right)\triangleq\|\mathbf{A}-\mu\left(\mathbf{A}\right)\|_{\text{F}}/\sqrt{N-1}$. Along the paper, we use the Normalized Cross-Correlation (NCC) between two matrices $\mathbf{A}$ and $\mathbf{B}$, defined as:
\begin{equation}
    \rho\left(\mathbf{A},\mathbf{B}\right)\triangleq\frac{\langle\mathbf{A}-\mu(\mathbf{A}),\mathbf{B}-\mu(\mathbf{B})\rangle_{\text{F}}}{\|\mathbf{A}-\mu(\mathbf{A})\|_{\text{F}}\|\mathbf{B}-\mu(\mathbf{B})\|_{\text{F}}}.
        \label{eq:NCC}
\end{equation}

\section{Preliminaries}
\label{sec:preliminaries}

This section outlines the key operations involved in PRNU-based camera source verification and explains Apple's portrait shooting mode, highlighting its impact on the PRNU.

\subsection{PRNU-based Camera Source Verification}
\label{subsec:PRNU_matching}

Luk{\'a}š \emph{et al.} \cite{LUKAS_2005} pioneered the use of camera sensor noise patterns, particularly the PRNU, to identify the specific camera that captured an image. To understand how the PRNU, which arises from tiny imperfections in the camera sensor, is used for camera source verification, we first consider the assumed sensor output model. For a single-channel image, denoted by a matrix $\mathbf{Y}$, the sensor output model can be approximated by the first two terms of its Taylor series~\cite{CHEN_2008}, as
\begin{equation}
        \mathbf{Y} = \left(\mathbf{1}+\mathbf{K}\right)\circ\mathbf{X}+\mathbf{\Theta},
        \label{eq:Y}
\end{equation}
where $\mathbf{K}$ is the PRNU signal, $\mathbf{X}$ is the incident light intensity and $\mathbf{\Theta}$ represents other noise sources.

\subsubsection{Baseline PRNU fingerprint extraction}
\label{subsubsec:PRNU_extraction}

Based on the model in \eqref{eq:Y}, which links the sensor output to the PRNU, a standard procedure has been established to estimate the PRNU of a given camera. This involves capturing a set of $L$ native-resolution images, i.e., $\left\{\mathbf{Y}_{l}\right\}_{l=1}^{L}$. Since the incident light intensity $\mathbf{X}$ in \eqref{eq:Y} is unknown, it is approximated via a denoising operation $F(\cdot)$ (here, we use the filter from \cite{MIHCAK_1999}), yielding $\hat{\mathbf{X}}=F(\mathbf{Y})$. The resulting residue $\mathbf{W}_l=\mathbf{Y}_{l}-F(\mathbf{Y}_{l})$ for each image $\mathbf{Y}_{l}$ is used to estimate the PRNU through the Maximum Likelihood Estimator (MLE) from \cite{CHEN_2008}:
\begin{equation}
        \hat{\mathbf{K}}=\left(\sum_{l=1}^{L}\mathbf{W}_l\circ\mathbf{Y}_l\right)\circ\left(\sum_{l=1}^{L}\mathbf{Y}_l\circ\mathbf{Y}_l\right)^{\circ-1}.
        \label{eq:hat_K}
\end{equation}
This procedure assumes all images $\mathbf{Y}_l$ are spatially aligned, with no geometric transformations and with their underlying PRNU patterns consistent pixel by pixel. Use of flat-field images is recommended to minimize content leakage \cite{FERNANDEZ-MENDUINHA-2021}.

\subsubsection{Baseline PRNU detection}
\label{subsubsec:PRNU_detection}

Given a test image $\mathbf{Y}_t$ with residue $\mathbf{W}_t=\mathbf{Y}_t-F(\mathbf{Y}_t)$ and a PRNU estimate $\hat{\mathbf{K}}$, the following hypothesis testing problem is formulated:
\begin{equation*}
        \begin{aligned}
                \mathcal{H}_0&: \text{$\mathbf{W}_t$ does not contain the PRNU $\mathbf{K}$,}\\
                \mathcal{H}_1&: \text{$\mathbf{W}_t$ contains the PRNU $\mathbf{K}$.}
        \end{aligned}
\end{equation*}
If $\mathcal{H}_1$ holds, the test image $\mathbf{Y}_t$ likely originates from the camera with PRNU $\mathbf{K}$. In image forensics, this decision is typically made using the Peak-to-Correlation Energy (PCE) statistic, which was validated in \cite{GOLJAN_2009} on a dataset of over a million images from 6,896 cameras (150 models). A threshold of 60 yielded a false alarm rate of $10^{-5}$ and a detection rate of $0.9762$. The PCE is calculated over multiple shifts to account for potential sensor misalignment. However, assuming spatial alignment and no cropping, we adopt a less computationally intensive alternative: the similarity measure $\eta$ inspired by the one proposed in \cite{PEREZGONZALEZ_2016}, defined in terms of the NCC in \eqref{eq:NCC} as:
\begin{equation}
        \eta\triangleq N\cdot\text{ssq}\left(\rho(\mathbf{W}_t,\hat{\mathbf{K}}\circ\mathbf{Y}_t)\right),
        \label{eq:eta}
\end{equation}
where the signed-squared function $\text{ssq}(\cdot)$ is defined as $\text{ssq}(x)\triangleq\text{sgn}(x)\cdot x^2$, and $\text{sgn}(\cdot)$ is a sign function that returns $-1$, $+1$, or $0$ depending on whether the input is negative, positive, or zero, respectively. Using this similarity measure, the hypothesis test can be performed by evaluating:
\begin{equation*}
        \eta\underset{\mathcal{H}_0}{\overset{\mathcal{H}_1}{\gtrless}}\tau,
\end{equation*}
where $\tau$ is a fixed threshold determined by the desired false positive probability. While $\eta$ differs from the PCE, their values are generally comparable, with $\eta$ having the added benefit of a significantly lower computational cost (cf. \cite{PEREZGONZALEZ_2016}). Unlike the standard PCE, which computes the ratio between the correlation peak (searched across all possible shifts to account for spatial misalignments) and the average correlation energy across all possible shifts, excluding a small neighborhood around the peak (commonly set to a size of $11\times11$), $\eta$ assumes spatial alignment and directly measures the NCC-based similarity. This simplification eliminates the need for exhaustive spatial searches, offering substantial efficiency gains while preserving comparable discriminative performance for aligned, uncropped images, which is the case in our experimental setup.

\begin{figure}[t]
        \centering
        \includegraphics[width=\linewidth]{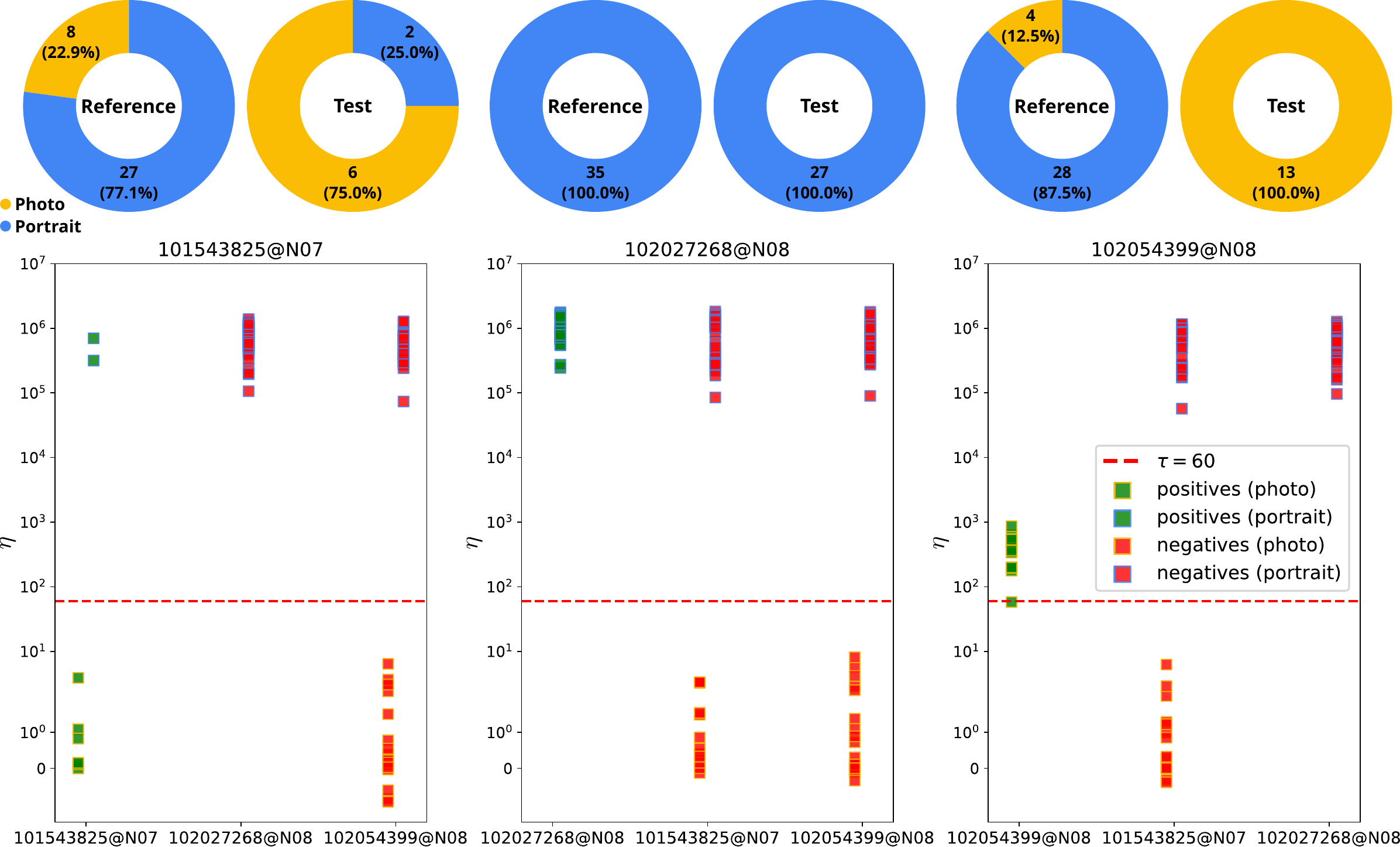}
        \caption{Values of $\eta$ for the 3 iPhone 11 Pro users with PRNU collisions in \cite{IULIANI_2021}. From left to right, each user's PRNU fingerprint is tested against the other two. The upper pie charts show the distribution of Reference and Test images per user, indicating the proportion of \texttt{Portrait} and \texttt{Photo} images. Note that all portrait samples, whether positive or negative, exhibit $\eta$ values well above the threshold separating the two classes, indicating that Apple's portrait mode increases false positives by introducing a distinct pattern that correlates more strongly than the device PRNU.}
        \label{fig:baseline_results_FORLAB}
\end{figure}

Our baseline PRNU matching approach, while using a slightly different PCE metric than Iuliani \emph{et al.} in \cite{IULIANI_2021}, successfully reproduces their results for the 3 iPhone 11 Pro users that yield PRNU collisions, as presented in Fig.~\ref{fig:baseline_results_FORLAB}.\footnote{To display both positive and negative values on a logarithmic scale, we use a symmetric logarithmic representation (concretely, \texttt{symlog} from \text{matplotlib}).} Iuliani \emph{et al.} suggest that images captured in Apple's portrait mode might contribute to user/device mismatches, consistent with findings from prior studies \cite{BARACCHI_2021} and \cite{ALBISANI_2021}. To further investigate this hypothesis, we analyze the $\eta$ values obtained for images taken in \texttt{Portrait} and \texttt{Photo} shooting modes, which are depicted in Fig.~\ref{fig:baseline_results_FORLAB} using blue and yellow colors, respectively.

Recall first that in the dataset from \cite{IULIANI_2021}, each user/device has ``Reference'' and ``Test'' image sets, with PRNU fingerprints extracted solely from the Reference set and positive samples taken from the corresponding Test set. For cross-user tests, all images from both sets are used. Notably, user \texttt{102027268@N08} has only portrait images, and false positives for this user occur exclusively in that mode (see middle plot of Fig. \ref{fig:baseline_results_FORLAB}). This suggests that Apple's portrait mode embeds a distinct pattern (here referred to as SDNP) that correlates more strongly than the PRNU, consistent with prior findings \cite{BARACCHI_2021}. As all Reference sets contain portrait images, PRNU fingerprints are influenced by the SDNP, leading to false positives when testing portrait images. Conversely, correct rejections occur only in standard \texttt{Photo} mode, where the SDNP is absent. A detailed analysis of the other two users is provided in the technical report \cite[Sect.~1]{SUP_MATERIAL_2025}. These results demonstrate that adjusting the detection threshold alone cannot resolve PRNU-based verification issues introduced by computational imaging techniques like the portrait mode.

In Sect.~\ref{subsec:Iuliani_2021_comparison}, these baseline results will be revisited after fully characterizing Apple's SDNP, beginning with the next section, which clarifies its origin.

\subsection{Apple Portrait Mode Description}
\label{subsec:apple_portrait_mode}

Standard smartphone cameras, due to their short focal lengths and small apertures, naturally produce images with a large depth of field, keeping most elements in focus. This limits their ability to achieve the aesthetically pleasing background blur, or bokeh, seen in professional photography. However, advances in computational photography have enabled software-based solutions that simulate a shallow depth of field. HTC pioneered this approach in 2014 with their ``Portrait Mode,'' and Apple popularized it two years later in the iPhone 7 Plus. Google also contributed by introducing innovative software-based depth sensing techniques rather than dual-camera hardware \cite{LEVOY_2018}. Today, most smartphone manufacturers offer a portrait mode, which introduces new forensic challenges for camera source attribution, as discussed in \cite{ALBISANI_2021}.

Apple's portrait mode implementation remains proprietary, but previous research by Baracchi \emph{et al.} \cite{BARACCHI_2021} has shed light on its behavior. Since then, the technology has evolved, prompting us to update the current understanding based on the latest iOS versions, focusing on iPhone models we tested directly (iPhone 16, 15, and 12 mini).\footnote{This study excludes iPad devices, though recent models support front-camera portrait mode. Future work will explore this, expecting similar results.} Initially introduced with the iPhone 7 Plus, portrait mode is available on models from the iPhone 8 Plus and iPhone X onward, as well as the iPhone SE (2nd generation) and later. With the iPhone 15, Apple introduced the ability to capture a regular photo and later convert it into a portrait in the \texttt{Photos} app if a person, dog, or cat is detected. Some iPhone models offer multiple zoom options (e.g., 1x and 2x), while older models like the iPhone XR and iPhone SE (2nd generation) require face detection to enable portrait mode with the rear camera. On the iPhone 15, users can also adjust zoom by pinching the screen. The standard resolution for portrait mode photos is 12MP ($4032\times3024$) for rear cameras and 7MP ($3088\times2316$) for front cameras.\footnote{Some older models have a slightly different resolution: $3088\times2320$.} Starting with the iPhone 15 series, Apple introduced a new maximum resolution of 24MP ($5712\times4284$). Since iOS 11 (2017), Apple has supported saving images in the High Efficiency Image File (HEIF) format \cite{HEIF_ISO}, with JPEG remaining an option for broader compatibility.

\begin{figure}[t]
        \centering
        \includegraphics[width=0.98\linewidth]{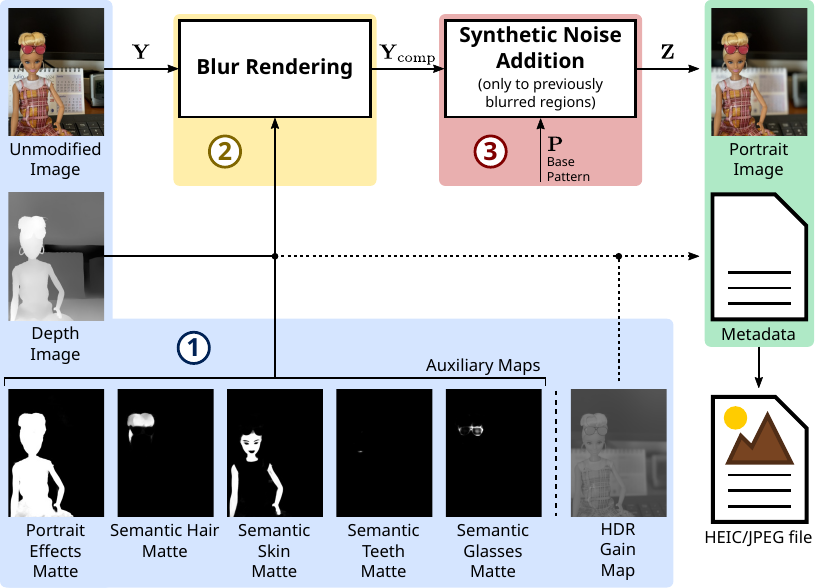}
        \caption{Apple's portrait mode block diagram highlighting the 3 main stages.}
        \label{fig:portrait_mode_diagram}
\end{figure}

To gain a general understanding of how software-based portrait modes function, the reader is referred to \cite{LEVOY_2018}. Based on our analysis of Apple's implementation, we identified three main stages, as depicted in the block diagram in Fig.~\ref{fig:portrait_mode_diagram}:

\subsubsection{Original Capture and Depth Map Generation}
\label{subsubsec:depth_map}

Apple's portrait mode first captures an unmodified image, denoted as $\mathbf{Y}$, which corresponds to the model in \eqref{eq:Y}. This original image can be retrieved from \texttt{iCloud Photos} or by disabling the portrait effect in the Photos app. Simultaneously, a depth map is generated using one of three methods \cite{APPLE_API_DEPTH}: (1) analyze disparities between multiple cameras (e.g., wide and telephoto); (2) apply machine learning to a single camera, particularly for human faces on select iPhone models; or (3) leverage specialized depth sensors like TrueDepth or LiDAR. However, these depth maps are lower resolution than the original image (see \cite[Tab.~1]{SUP_MATERIAL_2025}), limiting the precision of depth-based effects.

To enhance depth accuracy, iOS 12 introduced the depth-guided Portrait Effects Matte \cite{APPLE_API_MATTE}, a segmentation mask specifically designed for human subjects using a proprietary neural network. This matte refines subject-background separation, improving depth effects. In iOS 13, Apple extended semantic mattes to include details like hair, skin, teeth, and glasses (as shown in Fig.~\ref{fig:portrait_mode_diagram}).\footnote{While Fig.~\ref{fig:portrait_mode_diagram} includes an HDR gain map for reference, we found no direct connection to portrait mode and do not discuss it further in this paper.} These mattes, along with the depth map, are embedded in the image file and can be accessed using the open-source \texttt{libheif} library \cite{LIB_HEIF}. To the best of our knowledge, Apple employs a proprietary algorithm to integrate the depth map with auxiliary mattes, determining which areas will be blurred in the subsequent stage.

\subsubsection{Blur Rendering}
\label{subsubsec:blur_rendering}

Apple patents, such as \cite{BISHOP_2019}, describe potential methods for generating images with varying background blur levels, as documented in \cite{BARACCHI_2021}. Using the depth data from the previous stage, the original image $\mathbf{Y}$ is re-rendered with different degrees of blur applied based on pixel depth, the simulated aperture, and the 
relative position to the focal plane. The focal plane, which determines which pixels remain sharp, can be selected in the viewfinder, while the simulated aperture is adjustable via the \texttt{Depth Control} button, ranging from f/1.4 to f/16. The resulting composite image, denoted as $\mathbf{Y}_{\text{comp}}$, is expressed as:
\begin{equation}
        \mathbf{Y}_{\text{comp}} = \mathbf{M}_{(\text{blur})}^\prime\circ\mathbf{Y}+\mathbf{M}_{(\text{blur})}\circ \mathbf{Y}^\prime,
        \label{eq:Ycomp}
\end{equation}
where $\mathbf{M}_{(\text{blur})}$ is a binary mask with $1$'s in pixel positions where the blur has been applied and $0$ elsewhere, $\mathbf{M}^\prime_{(\text{blur})}$ is its logical negation, and $\mathbf{Y}^\prime\triangleq F_{\text{blur}}^{\text{f-stop}}(\mathbf{Y})$ represents the blurred version of $\mathbf{Y}$ based on the simulated aperture defined by the f-stop number. While the specifics of the blur rendering process are beyond the scope of this paper, different smartphone manufacturers define the shape of defocused areas at this stage (with some, like Samsung, allowing post-capture shape adjustments). Apple appears to use a circular bokeh effect.

At this stage, the composite image $\mathbf{Y}_{\text{comp}}$ may exhibit inconsistencies in noise levels, as applying algorithmic blur reduces natural noise in defocused regions compared to sharp areas. To achieve a more realistic depth-of-field effect, synthetic noise must be added, as discussed in the next section.

\subsubsection{Synthetic Noise Addition}
\label{subsubsec:synthetic_noise_addition}

The final step in generating the portrait image involves adding synthetic noise to the blurred regions of $\mathbf{Y}_{\text{comp}}$ (i.e., $\mathbf{Y}^\prime$) to create a realistic optical blur and minimize artifacts at the transitions between sharp and blurred areas (cf. \cite[Sect.~5.3]{LEVOY_2018}). For iPhone devices, we model the final portrait image $\mathbf{Z}$ as:
\begin{align}
        \mathbf{Z} &= \mathbf{Y}_{\text{comp}} + \mathbf{M}_{(\text{blur})}\circ\left( \gamma_{\text{ISO}}\cdot G\left(\mathbf{Y}^\prime\right)\circ\mathbf{P}+\mathbf{\Phi}\right)\nonumber\\
        &=\mathbf{M}^\prime_{(\text{blur})}\mspace{-3mu}\circ\mspace{-3mu}\mathbf{Y}\mspace{-2mu}+\mspace{-2mu}\mathbf{M}_{(\text{blur})}\mspace{-3mu}\circ\mspace{-3mu}\left(\mathbf{Y}^\prime\mspace{-2mu}+\mspace{-2mu}\gamma_{\text{ISO}}\cdot G\left(\mathbf{Y}^\prime\right)\mspace{-2mu}\circ\mspace{-2mu}\mathbf{P}\mspace{-2mu}+\mspace{-2mu}\mathbf{\Phi}\right),
        \label{eq:Z}
\end{align}
where \eqref{eq:Ycomp} is applied in the second equality. This model implies that unblurred regions of $\mathbf{Z}$ retain the sensor output characteristics defined in \eqref{eq:Y}, preserving the PRNU, while blurred regions do not reliably exhibit PRNU traces due to low-pass filtering and the addition of Apple's SDNP. The term $\mathbf{\Phi}$ is used to model various sources of noise resulting from full-frame processes in blurred areas such as compression, clipping, and other operations (including remaining traces of the original PRNU).

The added SDNP, modeled as $\mathbf{N}\triangleq\gamma_{\text{ISO}}\cdot G\left(\mathbf{Y}^\prime\right)\circ\mathbf{P}$, is a content-dependent pattern derived from a {\em Base Pattern} (BP) $\mathbf{P}$. Although this BP is fixed for a given iPhone and iOS version (see Sect.~\ref{sec:BP_variations} for details), we assume it can be modeled as a realization of a wide-sense stationary random process with mean $\mu_{\mathbf{P}}=0$ (as empirically confirmed in Sect.~\ref{subsec:BP_ext_stage_light_mono})  and standard deviation (std) $\sigma_{\mathbf{P}}=1$. This unit-std assumption is made for convenience as any scaling of the std could be absorbed into either $\gamma_\text{ISO}$ or $G\left(\mathbf{Y}^\prime\right)$.

This normalized BP is then adapted to the content of the captured scene by two scaling operations: 1) multiplication by an ISO-dependent scaling factor, $\gamma_{\text{ISO}}$ (further characterized in Sect.~\ref{subsec:gamma_ISO_estimation}); and 2) element-wise multiplication by a brightness-dependent matrix $G\left(\mathbf{Y}^\prime\right)$ with the same dimensions as $\mathbf{Y}^\prime$. We model the $(i,j)$th element of this matrix as  $\left[G\left(\mathbf{Y}^\prime\right)\right]_{i,j}=g\left(D_{i,j}\left(\mathbf{Y}^\prime\right)\right)$, where $g: \mathbb{R} \rightarrow \mathbb{R}$ is a function that will be determined in Sect.~\ref{subsec:G_estimation}, and we assume that operator $D_{i,j}(\mathbf{Y}^\prime)$ computes a local (possibly weighted) average of pixels of $\mathbf{Y}^\prime$ in a local neighborhood around $(i,j)$.

The base pattern $\mathbf{P}$, extracted as detailed in Sect.~\ref{sec:BP_extraction}, is the core component of the SDNP, acting as a fingerprint of Apple portrait images. Its presence not only explains the PRNU collisions observed in Apple portrait images (see Fig.~\ref{fig:baseline_results_FORLAB}), but also provides a basis for developing new forensic applications, as further detailed in Sect.~\ref{sec:BP_applications}. The next two sections validate this model and describe procedures for estimating $\mathbf{P}$.

\section{BP Extraction from Apple Portrait Images}
\label{sec:BP_extraction}

\begin{figure}[t]
        \centering
        \subfloat[]{\includegraphics[width=\linewidth]{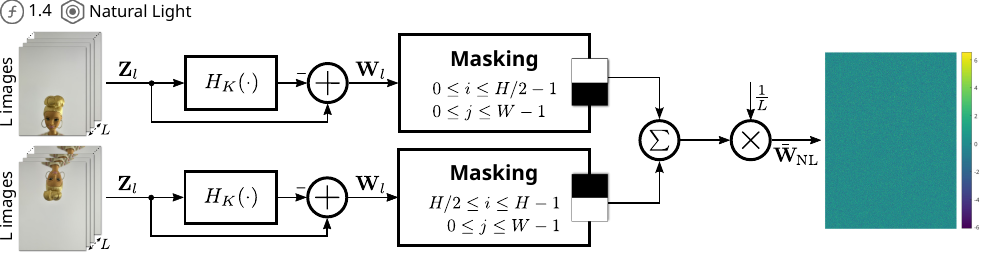}%
        \label{subfig:BP_ext_natural_light}}
        \\\vspace{0.15cm}
        \subfloat[]{\includegraphics[width=\linewidth]{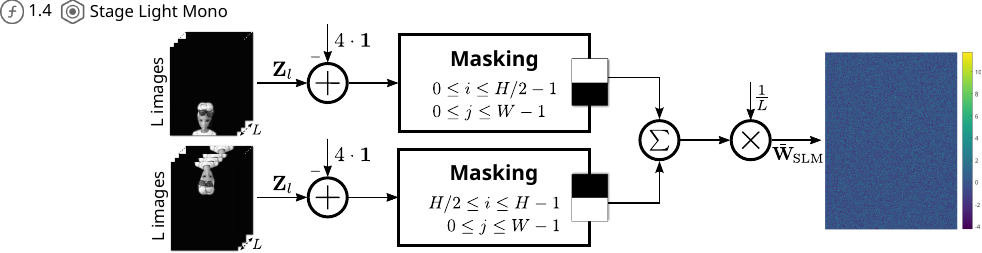}%
        \label{subfig:BP_ext_stage_light_mono}}
        \caption{Block diagram of the extraction process for the averaged residual matrix in each portrait lighting mode under study: $\bar{\mathbf{W}}_{\text{NL}}$ (a) and $\bar{\mathbf{W}}_{\text{SLM}}$ (b).}
        \label{fig:diff_scenes_BP_extraction}
\end{figure}

Following a PRNU-like approach, we outline a procedure to extract the BP fingerprint $\mathbf{P}$ from Apple portrait images. Unlike PRNU estimation, flat-field images are unsuitable here, since a foreground subject is always required. Instead, we use ``flat-background'' images with a uniform background to ensure a consistent BP application. Because a foreground subject is required, extracting the full-resolution BP involves capturing two scenes with alternating foreground and background regions: in the first scene, the subject is positioned in the lower third of the frame, ensuring a uniform background in the upper half; in the second scene, the subject is placed in the upper third, ensuring a uniform background in the lower half. Fig.~\ref{fig:diff_scenes_BP_extraction} illustrates this process for the two extraction methods discussed below.

\subsection{BP extraction with Portrait Lighting: Natural Light}
\label{subsec:BP_ext_natural_light}

For a given iPhone under investigation, this extraction process involves selecting the default \texttt{Natural Light} (NL) portrait effect and capturing $2L$ portrait images across the previously described scenes. This includes $L$ ``top'' images with a uniform background in the upper half and $L$ ``bottom'' images with a uniform background in the lower half. Each set is processed separately, and the results are aggregated to form an averaged residual matrix, $\bar{\mathbf{W}}_{\text{NL}}$, which serves as the basis for estimating the BP.

We first describe how to obtain the ``top'' half of $\bar{\mathbf{W}}_{\text{NL}}$, i.e., $\left(\bar{\mathbf{W}}_{\text{NL}}\right)_{i,j}$ for $0\leq i\leq H/2-1$ and $0\leq j\leq W-1$. For each of the $L$ ``top'' images, denoted as $\mathbf{Z}_l$ with $l=1,\dots,L$, a denoising filter is applied to the luminance component\footnote{The rationale for using the luminance component is detailed in Sect.~\ref{sec:estimation_BP_scaling_factors}.} to extract the noise residuals:
\begin{equation}
        \mathbf{W}_l = \mathbf{Z}_l-H_K\left(\mathbf{Z}_l\right),
        \label{eq:W_l}
\end{equation}
where $H_K(\cdot)$ represents the denoising operation. Unlike the denoising method used for PRNU extraction (i.e., \cite{MIHCAK_1999}), we employ a simple linear filter that convolves $\mathbf{Z}_l$  with a normalized $K\times K$ box kernel, due to the uniform background in the region of interest of the captured images. The choice of $K$ for BP extraction from these ``flat-background'' images is discussed in the technical report \cite[Sect.~2.1]{SUP_MATERIAL_2025}. The top half of $\bar{\mathbf{W}}_{\text{NL}}$ is then obtained by averaging the corresponding portions of the residuals as follows:
\begin{equation}
    \bar{\mathbf{W}}_{\text{NL}}\triangleq\frac{1}{L}\sum_{l=1}^{L} \mathbf{W}_l.
    \label{eq:bar_W}
\end{equation}
Similarly, after denoising the $L$ ``bottom'' images as in \eqref{eq:W_l}, the ``bottom'' half of $\bar{\mathbf{W}}_{\text{NL}}$, specifically the elements $\left(\bar{\mathbf{W}}_{\text{NL}}\right)_{i,j}$ for $H/2\leq i\leq H-1$ and $0\leq j\leq W-1$, is obtained by averaging the corresponding bottom portions of the residuals, using the same averaging process as in \eqref{eq:bar_W}.

The preceding procedure for obtaining $\bar{\mathbf{W}}_{\text{NL}}$ (illustrated in Fig.~\ref{subfig:BP_ext_natural_light}) does not directly estimate $\mathbf{P}$ in \eqref{eq:Z} due to the scaling operators $\gamma_{\text{ISO}}$ and $G\left(\mathbf{Y}^\prime\right)$. However, the uniform nature of the blurred ``flat-background'' regions enables two key approximations, allowing us to estimate $\mathbf{P}$ up to a constant scaling factor. First, assuming both a zero-mean BP (discussed further below) and a zero-mean noise component $\mathbf{\Phi}$ in \eqref{eq:Z}, we can approximate $H_K\left(\mathbf{Z}_l\right)\approx\mathbf{Y}_l^\prime$ in these smooth regions. Second, the scaling function can be approximated as $G\left(\mathbf{Y}_l^\prime\right)\approx G\left(\mu\left(\mathbf{Y}_l^\prime\right)\cdot\mathbf{1}\right)$, i.e., $\left[G\left(\mathbf{Y}_l^\prime\right)\right]_{i,j} =g\left(D_{i,j}\left(\mathbf{Y}^\prime\right)\right)\approx g\left(\mu\left(\mathbf{Y}_l^\prime\right)\right), \forall i,j$. This approximation is a direct consequence of the assumed nature of operator $D_{i,j}(\cdot)$ and the fact that $\mathbf{Y}^\prime_l$ is approximately constant. 
Notice that for clarity and brevity, we will adopt the shorthand notation $y^\prime\triangleq\mu\left(\mathbf{Y}^\prime\right)$ (or similarly, $y_l^\prime\triangleq\mu\left(\mathbf{Y}_l^\prime\right)$), especially when $\mu(\cdot)$ is an argument of the function $g(\cdot)$. Consequently, for this scenario, \eqref{eq:W_l} simplifies to:
\begin{equation}
        \mathbf{W}_l = \gamma_{\text{ISO}}\cdot g(y_l^\prime)\cdot\mathbf{P}+\mathbf{\Psi}_l,
        \label{eq:w_blur}
\end{equation}
where $\mathbf{\Psi}_l$ comprises $\mathbf{\Phi}_l$ and errors from the denoising process. Provided that the same ISO value is used for all $2L$ captured images (i.e., $\gamma_{\text{ISO}}$ is constant), it follows from \eqref{eq:bar_W} and \eqref{eq:w_blur} that:
\begin{equation}
        \bar{\mathbf{W}}_{\text{NL}}=\gamma_{\text{ISO}}\cdot\bar{\lambda}_{\text{NL}}\cdot\mathbf{P}+\bar{\mathbf{\Psi}}_{\text{NL}},
        \label{eq:approx_bar_W_NL}
\end{equation}
where $\bar{\lambda}_{\text{NL}} \triangleq \frac{1}{L}\sum_{l=1}^{L}g(y_l^\prime)$ represents the average value of the scaling function across the set of images and $\bar{\mathbf{\Psi}}_{\text{NL}}\triangleq\frac{1}{L}\sum_{l=1}^{L}\mathbf{\Psi}_l$ is the estimation noise. To simplify the analysis, we will assume that the noise components $\mathbf{\Psi}_l$ correspond to independent realizations of a wide-sense stationary and ergodic process. One consequence of this assumption is that $\lim_{L \rightarrow \infty} \|\bar{\mathbf{\Psi}}_{\text{NL}}\|_{\text{F}} = 0$. Notice that in \eqref{eq:w_blur} and \eqref{eq:approx_bar_W_NL} we should write $\left(\mathbf{P}-\mu\left(\mathbf{P}\right)\right)$ instead of $\mathbf{P}$, because the denoising process we use (i.e., $H_K(\cdot)$) removes the mean of $\mathbf{P}$. However, as we will confirm in Sect.~\ref{subsec:BP_ext_stage_light_mono}, $\mu\left(\mathbf{P}\right)\approx0$, so it is reasonable and notationally simpler to keep \eqref{eq:w_blur} and \eqref{eq:approx_bar_W_NL} unchanged. From \eqref{eq:approx_bar_W_NL} and the assumption that $\mathbf{P}$ is a realization of a stochastic process with zero mean and unit std, it follows that the normalized averaged residual $\bar{\mathbf{W}}_{\text{NL}}$ yields an estimate of $\mathbf{P}$, that is, $\hat{\mathbf{P}}_{\text{NL}}=\bar{\mathbf{W}}_{\text{NL}}/\sigma\left(\bar{\mathbf{W}}_{\text{NL}}\right)$. Since the quality of $\hat{\mathbf{P}}_{\text{NL}}$ depends on a well-curated $\bar{\mathbf{W}}_{\text{NL}}$, we offer guidelines on background selection and capture settings in \cite[Sect.~2.2]{SUP_MATERIAL_2025}. 

\begin{figure}[t]
        \centering
        \subfloat[ISO 125, $L=110$]{\includegraphics[width=0.49\linewidth]{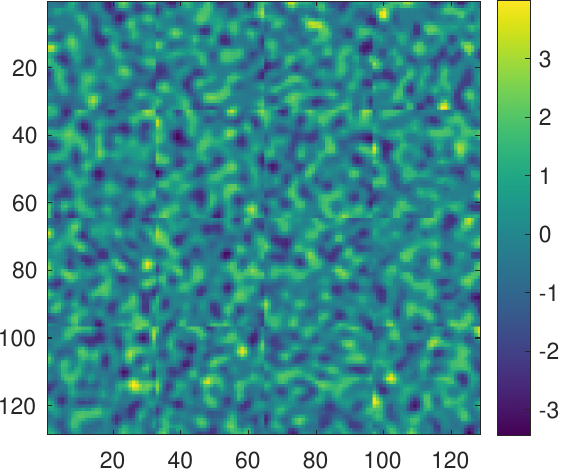}%
        \label{subfig:P_NL_128}}
        \hfill
        \subfloat[]{\includegraphics[width=0.49\linewidth]{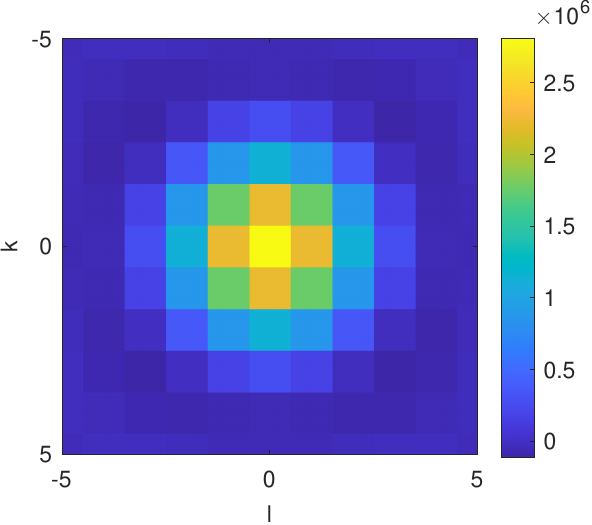}%
        \label{subfig:P_NL_autocorr}}
        \caption{Upper-left $128\times128$ patch of $\hat{\mathbf{P}}_{\text{NL}}$ estimated from 12MP HEIF images on the iPhone 15 (a). Autocorrelation of $\hat{\mathbf{P}}_{\text{NL}}$ limited to $k,l\in[-5,5]$ (b).}
        \label{fig:P_NL_patch_and_autocorr}
\end{figure}

As an illustrative example, the upper-left 128×128 patch of $ \hat{\mathbf{P}}_{\text{NL}} $ is shown in Fig.~\ref{subfig:P_NL_128}. Analyzing its properties further, the autocorrelation of $ \hat{\mathbf{P}}_{\text{NL}} $ (plotted in Fig.~\ref{subfig:P_NL_autocorr} for lags $ k,l\in[-5,5] $) indicates a colored, rather than white, noise process. This observation points towards underlying low-pass filtering, consistent with the analysis in \cite[Sect.~3.2]{SUP_MATERIAL_2025}. For additional details on BP characteristics, see \cite[Sect.~4.2]{SUP_MATERIAL_2025}. Finally, note that although a Minimum Mean Square Error (MMSE) estimate of $\mathbf{P}$ is conceivable, it would require knowledge of the second-order statistics of $\bar{\mathbf{\Psi}}_{\text{NL}}$, while offering only a performance gain that vanishes as $L$ increases. Its computation and analysis are left for future work.

\subsection{BP extraction with Portrait Lighting: Stage Light Mono}
\label{subsec:BP_ext_stage_light_mono}

The \texttt{Stage Light Mono} (SLM) portrait effect offers a convenient alternative for estimating Apple's BP from an iPhone under study. This effect simulates black-and-white stage lighting by isolating the subject in focus against a uniformly black background (see Fig.~\ref{subfig:BP_ext_stage_light_mono}). Its key advantage is that it removes the need for a physically uniform background, as depth information ensures a constant background luminance of 4 (i.e., $\mathbf{Y}^\prime=4\cdot\mathbf{1}$ in \eqref{eq:Z}) before adding the SDNP (an observation empirically validated in 
the technical report \cite[Sect.~3.1]{SUP_MATERIAL_2025}). However, this low luminance level causes values below zero being clipped, leading to saturation and thus information loss in the estimation of $\mathbf{P}$.

Using a methodology analogous to the NL-based BP extraction in Sect.~\ref{subsec:BP_ext_natural_light}, 
we capture $2L$ SLM portrait images, comprising $L$ ``top'' images with a black upper background and $L$ ``bottom'' images with a black lower background. Unlike the NL case, no filtering is needed to compute residuals, as $\mathbf{Y}^\prime$ is already known. The ``top'' half of the basis $\bar{\mathbf{W}}_{\text{SLM}}$ to estimate the BP, i.e., $\left(\bar{\mathbf{W}}_{\text{SLM}}\right)_{i,j}$ for $0\leq i\leq H/2-1$ and $0\leq j\leq W-1$, is obtained by subtracting the constant luminance matrix $4\cdot\mathbf{1}$ from each of the $L$ ``top'' images, $\mathbf{Z}_l$, and averaging the resulting matrices:
\begin{equation}
        \bar{\mathbf{W}}_{\text{SLM}}\triangleq\frac{1}{L}\sum_{l=1}^{L}\left(\mathbf{Z}_l - 4\cdot\mathbf{1}\right).
        \label{eq:bar_W_SLM}
\end{equation}
The ``bottom'' half of $\bar{\mathbf{W}}_{\text{SLM}}$, corresponding to elements $\left(\bar{\mathbf{W}}_{\text{SLM}}\right)_{i,j}$ for $H/2\leq i\leq H-1$ and $0\leq j\leq W-1$, is computed similarly by averaging the corresponding portions of the $L$ ``bottom'' images, using the same approach as in \eqref{eq:bar_W_SLM}.

In this scenario, since $\mathbf{Y}^\prime$ is constant, the scaling function $G\left(\mathbf{Y}^\prime\right)$ in \eqref{eq:Z} also yields a constant matrix. Thus, assuming a fixed ISO value is used across all $2L$ captured portrait images, our basis for estimating $\mathbf{P}$ is given, according to \eqref{eq:bar_W_SLM}, by: 
\begin{equation}
    \bar{\mathbf{W}}_{\text{SLM}} = \gamma_{\text{ISO}}\cdot\lambda_{\text{SLM}}\cdot\mathbf{P}+\bar{\mathbf{\Psi}}_{\text{SLM}},
    \label{eq:approx_bar_W_SLM}
\end{equation}
where $\lambda_{\text{SLM}} \triangleq g(4)$ is a constant factor and $\bar{\mathbf{\Psi}}_{\text{SLM}}\triangleq \frac{1}{L}\sum_{l=1}^{L}\mathbf{\Phi}_l$ represents the average noise component, including clipping effects from pixels saturated below 0. Unlike in the NL case, the noise components $\mathbf{\Phi}_l$ can no longer be assumed independent, as they originate from a constant scene in $\mathbf{Y}^\prime$ and the same ISO setting. Consequently, in the SLM case, $\|\bar{\mathbf{\Psi}}_{\text{SLM}}\|_{\text{F}}$ does not tend to zero as $L\rightarrow\infty$.

While, in principle, one could consider estimating $\mathbf{P}$ by normalizing $\bar{\mathbf{W}}_{\text{SLM}}$ as was done in the NL case (i.e., by computing $\bar{\mathbf{W}}_{\text{SLM}} / \sigma(\bar{\mathbf{W}}_{\text{SLM}})$), the fact that $\bar{\mathbf{\Psi}}_{\text{SLM}}$ does not vanish (and is significant relative to the term containing $\mathbf{P}$) would result in a considerably biased estimate. Instead, in the SLM case, it is preferable to estimate $\mathbf{P}$ by simply dividing $\bar{\mathbf{W}}_{\text{SLM}}$ in \eqref{eq:approx_bar_W_SLM} by $\gamma_{\text{ISO}} \cdot \lambda_{\text{SLM}}$, where  $\lambda_{\text{SLM}} = g(4)$. The next section discusses the estimation of $\gamma_{\text{ISO}}$ and $g(\cdot)$.

\section{Estimation of BP's Scaling Factors}
\label{sec:estimation_BP_scaling_factors}

We now focus on the scaling operators $\gamma_{\text{ISO}}$ and $G\left(\cdot\right)$ within the Apple portrait image model in \eqref{eq:Z}, which modulate the base pattern $\mathbf{P}$ according to scene characteristics (as detailed in Sect.~\ref{subsubsec:synthetic_noise_addition}). Since $G\left(\cdot\right)$ yields a constant matrix under the SLM lighting effect, we use SLM portrait images to estimate $\gamma_{\text{ISO}}$ (Sect.~\ref{subsec:gamma_ISO_estimation}). On the other hand, to determine $G\left(\cdot\right)$, which requires capturing a full range of luminance values, we rely on NL portrait images (Sect.~\ref{subsec:G_estimation}).

\subsection{Estimation of the ISO-Dependent Scaling Factor}
\label{subsec:gamma_ISO_estimation}

\begin{figure}[t]
    \centering
    \subfloat[ISO 50]{\includegraphics[width=2.225cm]{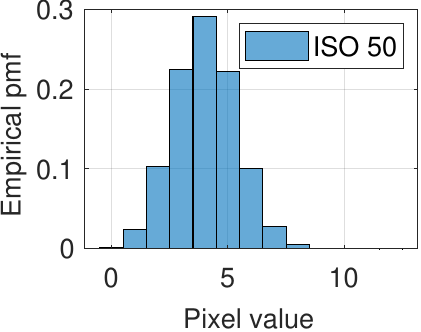}%
    \label{subfig:ISO50}}
    \hfill
    \subfloat[ISO 200]{\includegraphics[height=1.72cm]{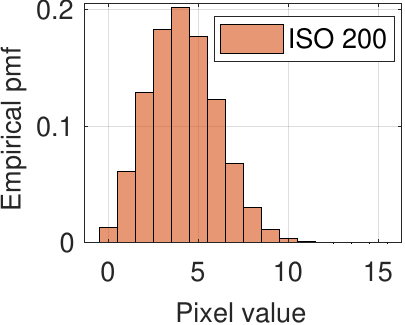}%
    \label{subfig:ISO200}}
    \hfill
    \subfloat[ISO 640]{\includegraphics[width=2.225cm]{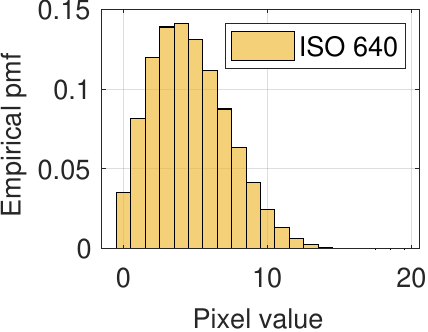}%
    \label{subfig:ISO640}}
    \hfill
    \subfloat[ISO 5000]{\includegraphics[width=2.225cm]{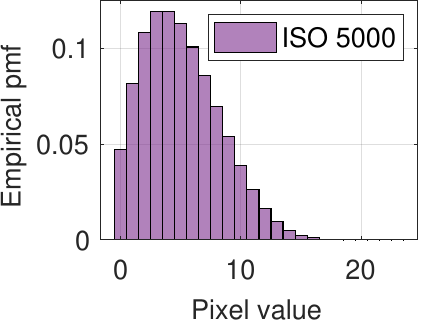}%
    \label{subfig:ISO5000}}
    \caption{Pixel value distribution in $\mathbf{Z}_{\text{SLM}}$ for different ISO values.}
    \label{fig:ISO_SLM}
\end{figure}

As noted in Sect.~\ref{subsec:BP_ext_stage_light_mono}, the SLM-based BP extraction ensures a constant brightness-dependent scaling factor, i.e., $G\left(\mathbf{Y}^\prime\right)=g(4)\cdot\mathbf{1}$. Since any gain can be absorbed into $\gamma_\text{ISO}$, we arbitrarily set $g(4)=1$, making $\lambda_{\text{SLM}}=1$ in \eqref{eq:approx_bar_W_SLM}. The effect of $\gamma_{\text{ISO}}$ can be noticed in Fig.~\ref{fig:ISO_SLM}, which illustrates the empirical probability mass function (pmf) of $\mathbf{Z}_{\text{SLM}}$ (a submatrix of the background region of an SLM portrait image containing the SDNP) for ISO values of 50, 200, 640, and 5000. As ISO increases, so does the scaling factor $\gamma_{\text{ISO}}$, resulting in a higher occurrence of saturated pixels.

The empirical pmfs in Fig.~\ref{fig:ISO_SLM} show that the right tail of the distribution, less affected by saturation, closely follows a Gaussian distribution. This approximation is confirmed by a Kolmogorov-Smirnov test 
(see \cite[Sect.~3.2]{SUP_MATERIAL_2025}). To exploit this observation, we conduct an exhaustive search to match the observed data distribution (i.e., the empirical pmf of $\mathbf{Z}_{\text{SLM}}$ for a given ISO) with synthetic Gaussian distributions. These distributions are generated using candidate mean and standard deviation values, simulating the effects of $\mu\left(\mathbf{P}\right)$ and $\gamma_{\text{ISO}}$, respectively. Specifically, we compare the observed pmf of $\mathbf{Z}_{\text{SLM}}$ for a given ISO value with the distribution (parameterized by $\mu_{Z}, \sigma_{Z}$) of the following random variable:
\begin{equation}
    Z= \max\left\{0,\text{round}\left(\mu_{Z} + \sigma_{Z} P\right)\right\}
    \label{eq:Z_synth}
\end{equation}
where $P$ is a unit-variance Gaussian random variable, and $\text{round}(\cdot)$ represents rounding to the nearest integer, mimicking the 8-bit depth of SLM portrait images. We do not explicitly model the compression that occurs in practice to avoid increasing model complexity and complicating the exhaustive search, which is already performed over two parameters.

To find estimates for parameters $\mu_Z, \sigma_Z$ we minimize the Kullback-Leibler Divergence (KLD)\footnote{We also tested the chi-square distance and obtained similar results.} between the empirical pmf $\mathbf{h}_{\text{SLM}}$ of the observations $\mathbf{Z}_{\text{SLM}}$ (for a given ISO value), and the empirical pmf $\mathbf{h}_Z$ of i.i.d. samples of $Z$. In our experiments, the estimation process involves exploring a predefined grid of candidate pairs $(\mu_Z,\sigma_Z)$ that minimize the KLD, i.e.,
\begin{equation*}
    (\hat{\mu}_Z,\hat{\sigma}_Z)\triangleq\arg\min_{(\mu_Z,\sigma_Z)\in\mathcal{M}\times\mathcal{S}}\text{KLD}\left(\mathbf{h}_{\text{SLM}},\mathbf{h}_Z\right),
\end{equation*}
where each candidate pair $(\mu_Z,\sigma_Z)\in\mathcal{M}\times\mathcal{S}$ is obtained by sampling the intervals $\mathcal{M}\triangleq\left[\mu\left(\mathbf{Z}_{\text{SLM}}\right)-1,\mu\left(\mathbf{Z}_{\text{SLM}}\right)+1\right]$ and $\mathcal{S}\triangleq\left[\sigma\left(\mathbf{Z}_{\text{SLM}}\right)-1,\sigma\left(\mathbf{Z}_{\text{SLM}}\right)+1\right]$ uniformly, with a step size of $0.1$. Following \eqref{eq:Z} for the particular case $\mathbf{Y}^\prime=4\cdot
\mathbf{1}$ and the assumption $g(4)=1$, we can model the blurred samples of the SLM image prior to quantization and clipping as $4 \cdot \mathbf{1}+\gamma_{\text{ISO}} \cdot \mathbf{P}$. 
On the other hand, we have just shown that these samples are well modeled by a Gaussian distribution with mean $\hat{\mu}_Z$ and std $\hat{\sigma}_Z$. Therefore, by simple identification, we can infer that $\hat {\gamma}_{\text{ISO}}=\hat{\sigma}_Z/\sigma_{\mathbf{P}}=\hat{\sigma}_Z$ (given our assumption that $\sigma_{\mathbf{P}}=1$) and $\hat{\mu}_{\mathbf{P}}=(\hat{\mu}_Z-4)/\hat{\gamma}_\text{ISO}$.

\begin{figure}[t]
        \centering
        \subfloat[]{\includegraphics[width=2.225cm]{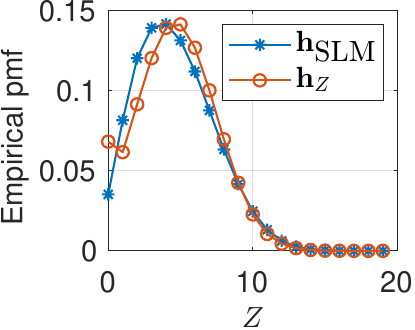}%
        \label{subfig:h_SLM_vs_h_Z}}
        \subfloat[]{\includegraphics[width=2.225cm]{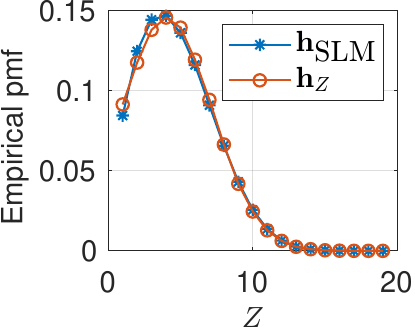}%
        \label{subfig:h_SLM_vs_h_Z_Z_lt_0}}
        \subfloat[]{\includegraphics[height=1.76cm]{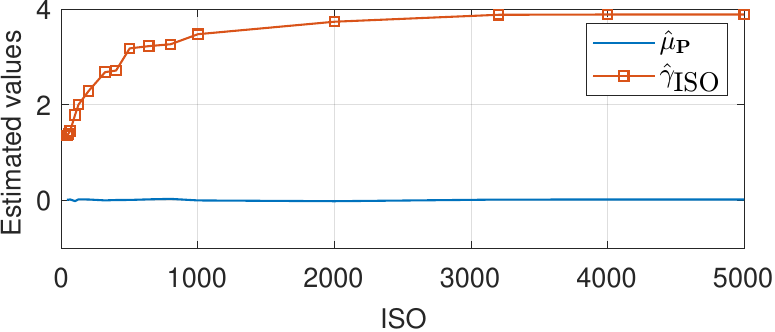}%
        \label{subfig:hat_mu_hat_gamma_ISO}}
        \caption{First two panels compare the empirical pmfs $\mathbf{h}_{\text{SLM}}$  and $\mathbf{h}_Z$ at ISO 500, for $Z\geq0$ (a) and $Z>0$ (b). The right panel (c) illustrates the estimated values of $\hat{\mu}_{\mathbf{P}}$ and $\hat{\gamma}_{\text{ISO}}$ across different ISO settings.}
        \label{fig:gamma_ISO_estimation}
\end{figure}

Lacking a suitable model for the compression operation, our initial estimate was biased by the peak at $Z=0$ due to clipping, as shown in Fig.~\ref{subfig:h_SLM_vs_h_Z} for the particular case of ISO 500. Compression would typically spread these values across neighboring bins. To reduce this bias, we excluded zero-valued pixels from both $\mathbf{Z}_{\text{SLM}}$ and the synthetic variable $Z$ when computing the KLD. This adjustment, illustrated in Fig.~\ref{subfig:h_SLM_vs_h_Z_Z_lt_0}, significantly improved the alignment of empirical pmfs and yielded more reliable estimates. Additional details on the exhaustive search using KLD and Chi-square distance are in 
\cite[Sect.~3.3]{SUP_MATERIAL_2025}.

Processing images at different ISO values yielded the results in Fig.~\ref{subfig:hat_mu_hat_gamma_ISO}, showing that the mean of $\mathbf{P}$ remains close to zero regardless of ISO. On the other hand, the scaling factor $\gamma_{\text{ISO}}$ increases with ISO but remains asymptotically bounded. A table listing the estimated $\hat{\gamma}_{\text{ISO}}$ values for various ISO settings is available in \cite[Tab.~3]{SUP_MATERIAL_2025}. With the knowledge of $\hat{\gamma}_{\text{ISO}}$, obtained under the assumption that $\lambda_{\text{SLM}}=1$, we can now estimate the BP from SLM portrait images as described at the end of Sect.~\ref{subsec:BP_ext_stage_light_mono}, resulting in $\hat{\mathbf{P}}_{\text{SLM}}=\frac{1}{\hat{\gamma}_{\text{ISO}}}\bar{\mathbf{W}}_{\text{SLM}}$.

\subsection{Estimation of the Brightness-Dependent Scaling Function}
\label{subsec:G_estimation}

With the ISO-dependent scaling factor $\gamma_{\text{ISO}}$ characterized, we now turn to the brightness-dependent scaling function $G\left(\mathbf{Y}^\prime\right)$. As mentioned previously, to capture the full luminance range (0 to 255 for 8-bit depth), we use NL portrait images with uniform patches. This allows us to approximate the scaling function as $G\left(\mathbf{Y}^\prime\right)\approx G\left(\mu\left(\mathbf{Y}^\prime\right)\cdot\mathbf{1}\right)$, where $\left[G\left(\mathbf{Y}^\prime\right)\right]_{i,j} \approx g(y^\prime), \forall i,j$ (recall that $y^\prime \triangleq \mu(\mathbf{Y}^\prime)$). Thus, we characterize $G\left(\mathbf{Y}^\prime\right)$ through its element-wise form $g(y^\prime)$.

\begin{figure}[t]
        \centering
        \subfloat[ISO 400]{\includegraphics[width=0.24\linewidth]{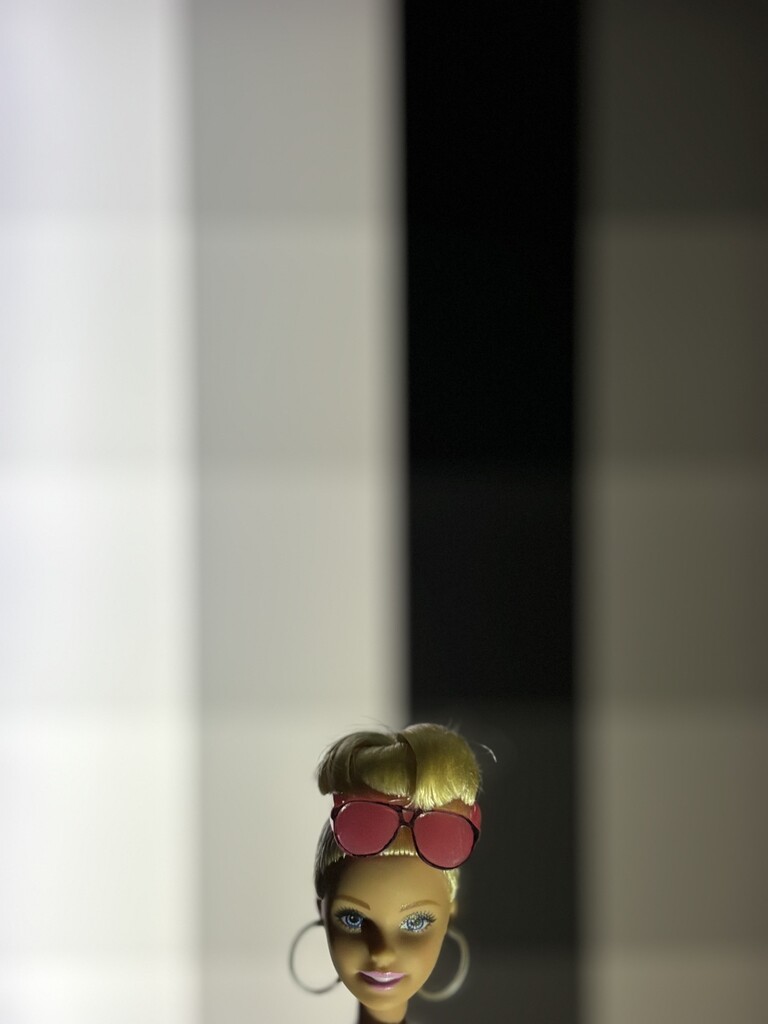}
        \hfill
        \includegraphics[width=0.24\linewidth]{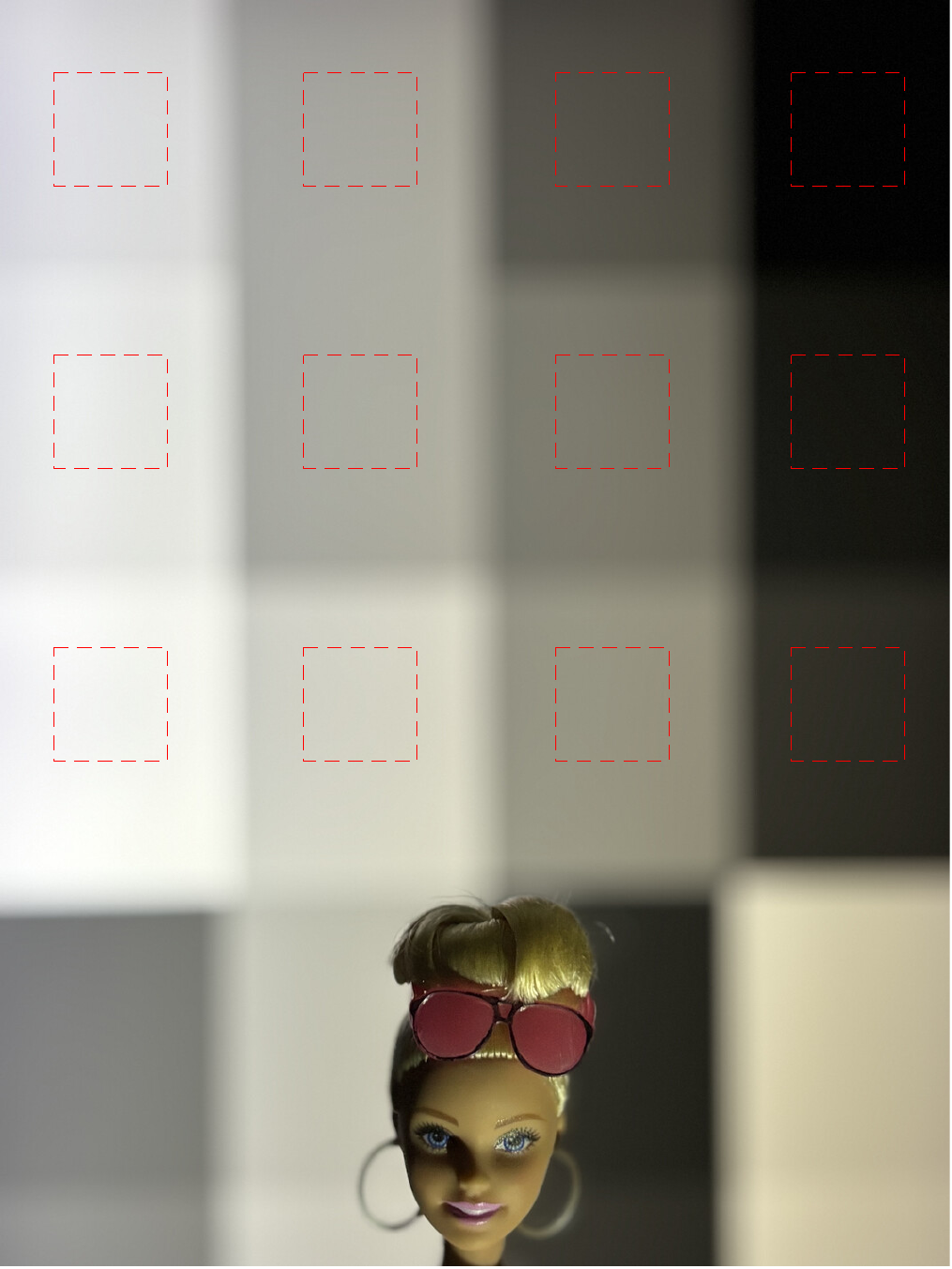}%
        \label{subfig:colorchecker_ISO_400}}
        \hfill
        \subfloat[ISO 125]{\includegraphics[width=0.24\linewidth]{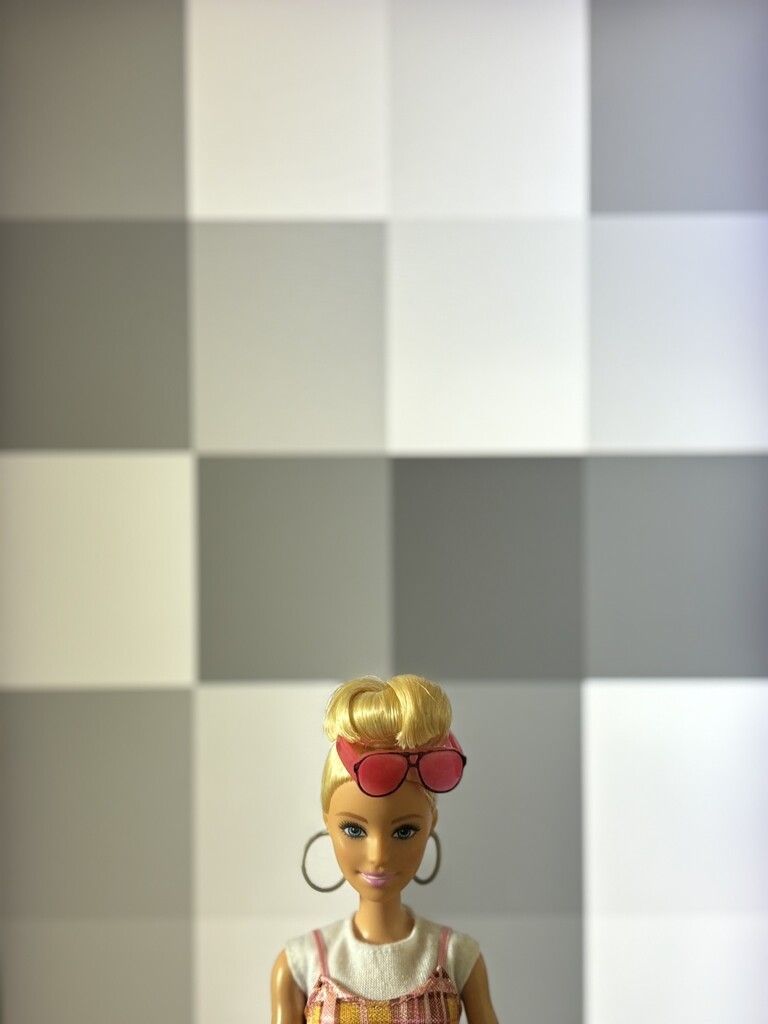}
        \hfill
        \includegraphics[width=0.24\linewidth]{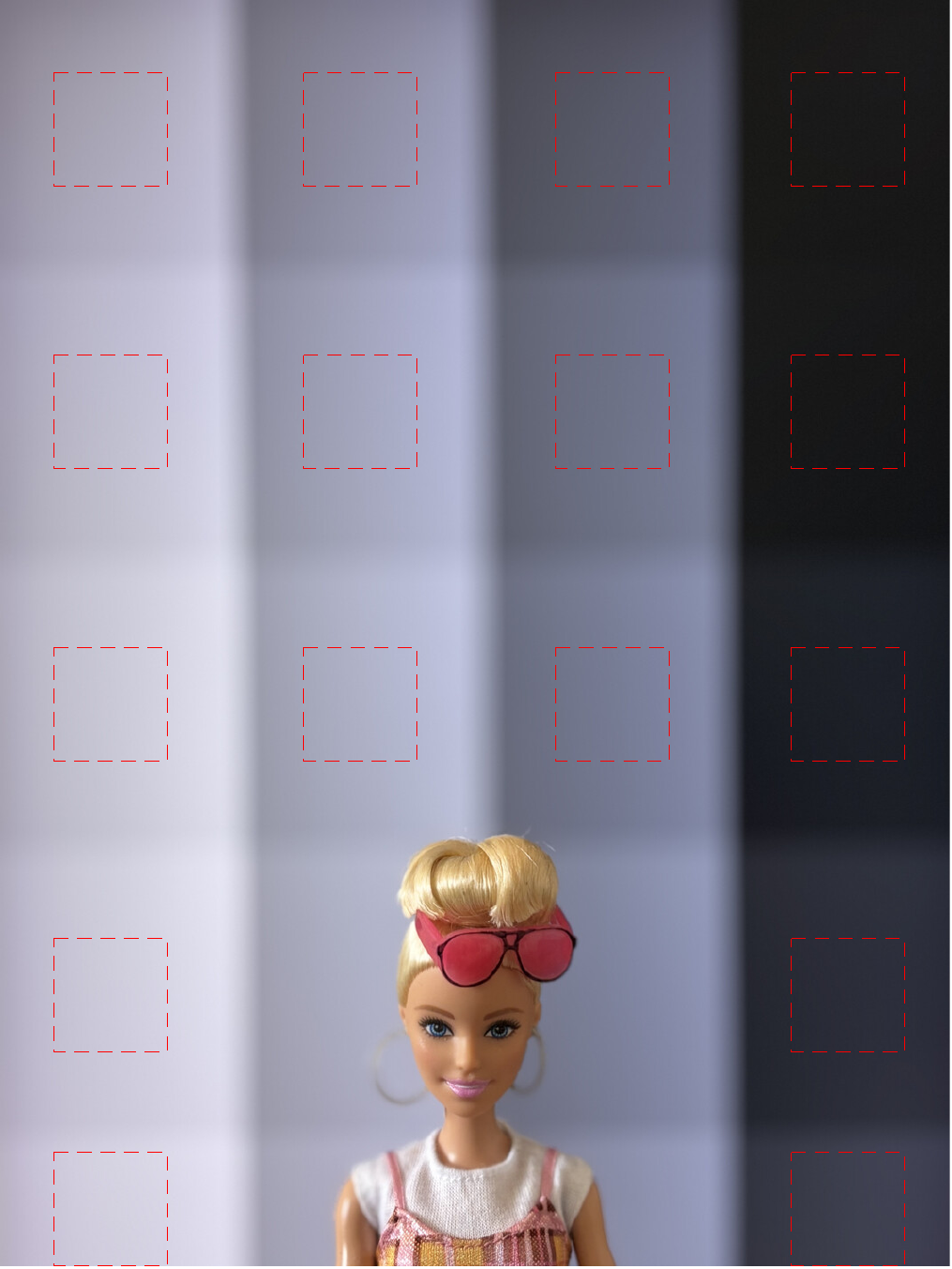}%
        \label{subfig:colorchecker_ISO_125}}
        \caption{Examples of portrait images of the ColorChecker scene captured at full resolution (24MP). The left panel (a) displays two samples from a set of $T=86$ images taken at ISO 400, while the right panel (b) shows two samples from a set of $T=64$ images taken at ISO 125. The processed $B\times B$ blocks ($B=512$) are highlighted in red in the rightmost image of each case.}
        \label{fig:colorchecker_gray}
\end{figure}

To estimate $g(\cdot)$, we collect $M$ flat-background patches of size $B \times B$ containing the SDNP from each 
of $T$ images captured at the same ISO, yielding a total of $M \cdot T$ patches. These patches are non-overlapping within each image, co-located across different images (i.e., they share the same coordinates across the $T$ images), and span different background luminance levels (illustrated in Fig.~\ref{fig:colorchecker_gray}). Let $\mathbf{Z}_{m,t}$ represent the $m$th patch from the $t$th image, with $\mathbf{Y}^\prime_{m,t}$ and $\mathbf{W}_{m,t}$ similarly defined. $\mathbf{P}_m$ denotes the $m$th patch of the BP, which is co-located with $\mathbf{Z}_{m,t}$ and of the same size. Since $z_{m,t} \triangleq \mu(\mathbf{Z}_{m,t})$ serves as an estimate of $y^\prime_{m,t} \triangleq \mu(\mathbf{Y}^\prime_{m,t})$, our goal is to obtain pairs $\left(z_{m,t}, \hat g(y'_{m,t})\right)$ as estimates of $\left(y'_{m,t}, g(y'_{m,t})\right)$. By carefully selecting the background levels $z_{m,t}$, we ensure that the entire range $[0,255]$ is well represented.

We will explain the proposed estimation procedure for the $m$th patch location, which can be straightforwardly generalized to the other patches. Recall that for this location we have a set of $T$ residuals of the form $\mathbf{W}_{m,t}=\gamma_{\text{ISO}}g(y_{m,t}^\prime)\mathbf{P}_m+\mathbf{\Psi}_{m,t}$, $t=1, \dots, T$. These equations can be more conveniently written by converting matrices to vector form using the $\text{vec}(\cdot)$ operator, which returns the vector obtained by stacking the columns of the input matrix. Let $\mathbf{w}_{m,t} \triangleq \text{vec}(\mathbf{W}_{m,t})$; $\mathbf{p}_{m} \triangleq \text{vec}(\mathbf{P}_{m})$; and $\boldsymbol{\psi}_{m,t} \triangleq \text{vec}(\mathbf{\Psi}_{m,t})$, where all these vectors have $B^2$ elements, then we can write $\mathbf{w}_{m,t}=\gamma_{\text{ISO}}g(y_{m,t}^\prime)\mathbf{p}_m+\boldsymbol{\psi}_{m,t}$, $t=1, \dots, T$. Under the assumptions detailed in Sect.~\ref{subsec:BP_ext_natural_light}, we have that $\mathbf{p}_m \sim \mathcal{N}(\mathbf{0}, \mathbf{I}_{B^2})$ and $\boldsymbol{\psi}_{m,t} \sim \mathcal{N}(\mathbf{0}, \sigma^2_{\psi_m} \cdot \mathbf{I}_{B^2})$, where $\sigma^2_{\psi_m}$ is unknown. Moreover, note that by our assumptions, vectors $\boldsymbol{\psi}_{m,t}$ and $\boldsymbol{\psi}_{m,u}$ are uncorrelated for any $t \neq u$. Finally, let $\mathbf{g}_m \triangleq [g(y'_{m,1}), \cdots , g(y'_{m,T})]^T \in \mathbb{R}^T$, $\mathbf{W}_m \triangleq [\mathbf{w}_{m,1}, \cdots, \mathbf{w}_{m,T}] \in \mathbb{R}^{B^2 \times T}$, and construct the corresponding Gram matrix of $\mathbf{W}_m$ as   
\begin{equation*}
    \mathbf{G}_m\triangleq\frac{1}{B^2}\mathbf{W}_m^T\mathbf{W}_m.
\end{equation*}
Then, from our model, it follows that its expected value is such that
\begin{equation*}
    \text{E}\{\mathbf{G}_m\}=\gamma_\text{ISO}^2\cdot \mathbf{g}_m \mathbf{g}_m^T + \sigma_{\psi_m}^2\cdot \mathbf{I}_{T}.
\end{equation*}
This indicates that the ``signal'' part of $\mathbf{W}_m$ lies in the one-dimensional subspace spanned by $\mathbf{g}_m$, and so this vector can be estimated via the eigendecomposition of $\mathbf{G}_m$.

Let $\lambda_1$ be the dominant eigenvalue of $\mathbf{G}_m$ and $\mathbf{v}_1$ its associated eigenvector. The noise variance can be estimated from the remaining eigenvalues as $\hat{\sigma}_{\psi_m}^2=\frac{1}{T-1}\sum_{j=2}^{T}\lambda_j$, and the signal strength as $s=\max\{\lambda_1-\hat{\sigma}_{\psi_m}^2,0\}$. Using these quantities, the vector of brightness-dependent scaling factors can be estimated as 
\begin{equation}
\hat{g}(y'_{m,t})=\frac{\sqrt{s}}{\gamma_{\text{ISO}}} {|v_{1,t}|}, \quad t=1, \dots, T.
\label{eq:norm_corr}
\end{equation}
The removal of the noise variance from $\lambda_1$ leads to $\hat{g}(y'_{m,t})$ being an unbiased estimator. 
The estimation process just described is repeated for each $m=1, \dots, M$. This results in $M \cdot T$ pairs $(z_{m,t}, \hat{g}(y'_{m,t}))$.

\begin{figure*}[t]
        \centering
        \subfloat[]{\includegraphics[width=0.24\linewidth]{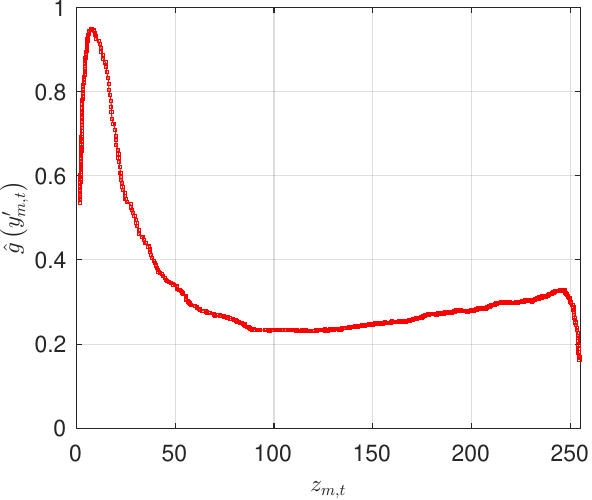}%
        \label{subfig:corr_vs_R}}
        \hfill
        \subfloat[]{\includegraphics[width=0.24\linewidth]{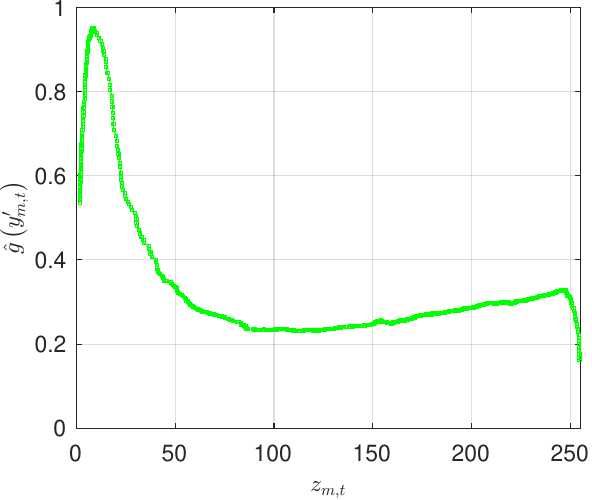}%
        \label{subfig:corr_vs_G}}
        \hfill
        \subfloat[]{\includegraphics[width=0.24\linewidth]{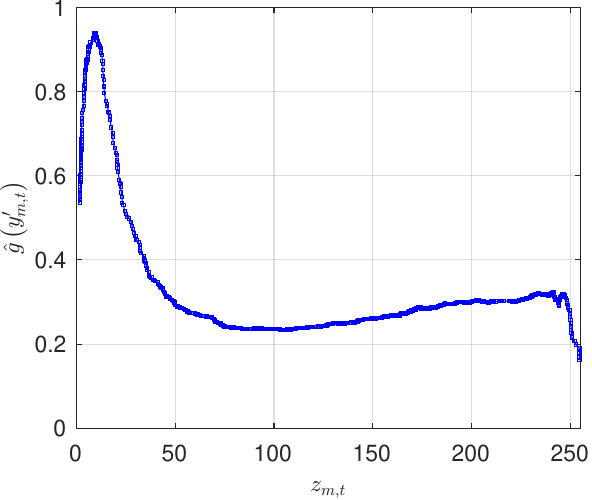}%
        \label{subfig:corr_vs_B}}
        \hfill
        \subfloat[]{\includegraphics[width=0.24\linewidth]{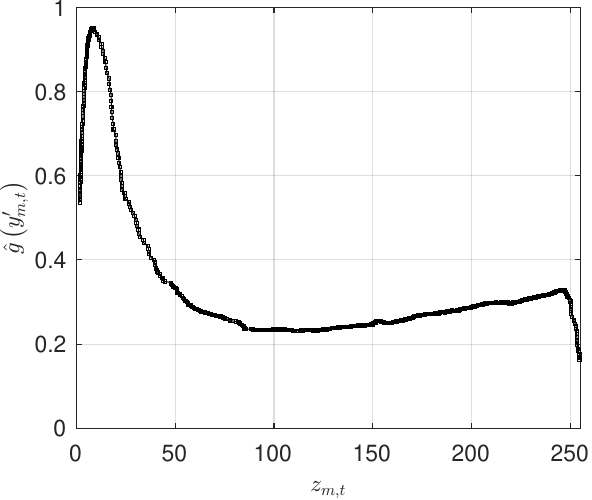}%
        \label{subfig:corr_vs_Y}}
        \hfill
        \caption{Evolution of the estimated brightness-dependent scaling function $\hat{g}(y^\prime_{m,t})$ as a function of $z_{m,t}$, based on co-located uniform patches captured at ISO 400. Results are presented for the R (a), G (b), and B (c) channels of the RGB colorspace, and the luminance component (d) of the YUV colorspace. All curves are smoothed using a 5-point moving average for each $z_{m,t}$.}
        \label{fig:correlation_vs_average_value}
\end{figure*}

We implemented this procedure by capturing NL portrait images of a ColorChecker-like board displayed on a screen behind the subject, creating uniform gray patches with the SDNP present in the background (see Fig.~\ref{fig:colorchecker_gray}). A total of $T=86$ images (captured at ISO 400) were taken using the iPhone 15 at 24MP resolution. From each image, we processed $M=12$ co-located $B\times B$ patches, with $B=512$ pixels (highlighted in red in the rightmost image of Fig.~\ref{subfig:colorchecker_ISO_400}), resulting in $M\cdot T=1032$ distinct patches in total. The resulting empirical pairs $(z_{m,t}, \hat{g}(y'_{m,t}))$ are plotted separately for the three RGB channels and the luminance component of the YUV color space in Fig.~\ref{fig:correlation_vs_average_value}. For better visualization, we plot a smoothed version of the values of $\hat{g}(y'_{m,t})$ using a 5-point moving average for each $z_{m,t}$.

The estimated scaling function behaves similarly across all four components, with the red and blue channels showing the most variability and luminance the least. Since RGB channels follow comparable trends, the BP is likely embedded primarily in the luminance component. Supporting this, the NCC between $\hat{\mathbf{P}}_{\text{NL}}$ (extracted from luminance) and each color channel was nearly 1: 0.990 (R), 0.995 (G), and 0.991 (B). While not conclusive, this justifies assuming that the BP is embedded in the luminance. Thus, we heretofore focus solely on the luminance component. If BP estimates are available, such as $\hat{\mathbf{P}}_{\text{NL}}$ or $\hat{\mathbf{P}}_{\text{SLM}}$, an alternative estimate of $g(\cdot)$ can be derived using a Least Squares (LS) approach. As detailed in Appendix~\ref{app:MMSE_g_y}, the resulting estimate is:
\begin{equation}
        \hat{g}\left(y'_{m,t}\right)=\frac{1}{\hat{\gamma}_{\text{ISO}}}\frac{\left\langle\mathbf{Z}_{m,t}, \hat{\mathbf{P}}_m\right\rangle_{\text{F}}-B^2 z_{m,t} \cdot \mu(\hat{\mathbf{P}}_m)}{\|\hat{\mathbf{P}}_m\|_{\text{F}}^2-B^2 \left(\mu(\hat{\mathbf{P}}_m)\right)^2},
        \label{eq:hat_g}
\end{equation}
where $\hat{\mathbf{P}}_m$ represents the $m$th $B\times B$ block of any available estimate of $\mathbf{P}$ (spatially aligned with $\mathbf{Z}_{m,t}$), and $\hat{\gamma}_{\text{ISO}}$ is the estimated scaling factor corresponding to the ISO value used to capture the observed blocks. Note that while we assume $\mu_{\mathbf{P}}=0$, this does not hold for certain BP estimates, such as $\hat{\mathbf{P}}_\text{SLM}$, where $\mu(\hat{\mathbf{P}}_\text{SLM})\neq0$ due to saturation effects. In Appendix~\ref{app:MMSE_g_y} we also derive the following estimate $\hat{y}^\prime_{m,t}=z_{m,t}-\hat{\gamma}_\text{ISO} \cdot \hat{g}(y'_{m,t}) \cdot \mu(\hat{\mathbf{P}}_m)$, which becomes $z_{m,t}$ when $\mu(\hat{\mathbf{P}}_m)=0$. 

\begin{figure}[t]
        \centering
        \subfloat[]{\includegraphics[width=0.5\linewidth]{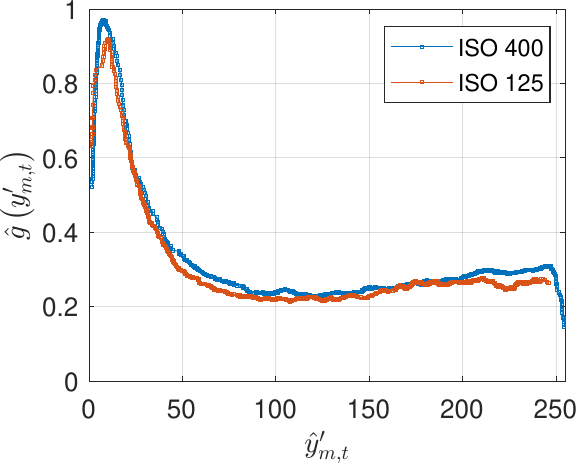}%
        \label{subfig:hat_g_y_diff_ISO}}    
        \hfill
        \subfloat[]{\includegraphics[width=0.5\linewidth]{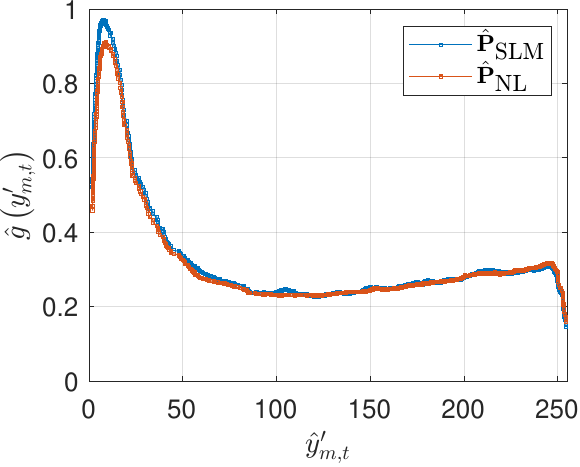}%
        \label{subfig:hat_g_diff_BP}}
        \caption{Estimated pairs $(\hat y'_{m,t}, \hat{g}(y'_{m,t}))$ using the LS approach. Panel (a) compares results for $\hat{\mathbf{P}}_{\text{SLM}}$ at ISO 400 and 125. Panel (b) contrasts $\hat{\mathbf{P}}_{\text{SLM}}$ and $\hat{\mathbf{P}}_{\text{NL}}$ at ISO 400. All curves are smoothed using a 5-point moving average.}
        \label{fig:hat_g_y}
\end{figure}

We plot the pairs $\left(\hat y'_{m,t}, \hat{g}\left(y'_{m,t}\right)\right)$ for $m=1,\dots,M$ and $t=1,\dots,T$ smoothed with a 5-point moving average as the final LS estimate.  Fig.~\ref{subfig:hat_g_y_diff_ISO} shows the resulting curves obtained using $\hat{\mathbf{P}}_{\text{SLM}}$ as estimator of $\mathbf{P}$ for two different ISO values. For ISO 400, we use the aforementioned 1032 distinct blocks from $M=12$ co-located blocks across $T=86$ ColorChecker portrait images. Similarly, for ISO 125, we use 1072 blocks captured from $T=67$ images and $M=16$ co-located blocks (highlighted in red in the rightmost image of Fig.~\ref{subfig:colorchecker_ISO_125}). Although the obtained curves are not identical, their alignment after applying the corresponding $\hat{\gamma}_{\text{ISO}}$ values for ISO 125 and ISO 400 supports the consistency of the model assumed in \eqref{eq:Z}. To compare the estimated scaling functions $\hat{g}\left(\cdot\right)$ derived from different BP estimates, we compute \eqref{eq:hat_g} using $\hat{\mathbf{P}}_{\text{SLM}}$ and $\hat{\mathbf{P}}_{\text{NL}}$ for the ISO 400 case. Fig.~\ref{subfig:hat_g_diff_BP} shows that the resulting curves match almost perfectly. Finally, the near-identical curves in Figs.~\ref{fig:correlation_vs_average_value} and \ref{fig:hat_g_y} (from various estimation methods and ISO values) demonstrate the convergence of different approaches and data to very similar brightness-dependent scaling function estimates, providing strong evidence for the model's validity. 

Although not explored here, the estimated function $\hat{g}(\cdot)$ could be used to emphasize the extracted BP, as proposed in \cite{FERNANDEZMENDUINA_2023} for the PRNU, but this is left for future work. Meanwhile, after characterizing Apple's SDNP by estimating its scaling parameters and underlying BP, we analyzed key factors affecting BP estimate quality, including the number of images $L$, scene brightness, portrait lighting modes, and ISO settings. Details of this analysis, including a study on the variability of $\hat{g}(\cdot)$ across different devices and iOS versions, are provided in \cite[Sect.~4]{SUP_MATERIAL_2025}.

\section{Analysis of Apple's BP Variations}
\label{sec:BP_variations}

\begin{figure}[t]
        \centering
        \subfloat[]{\includegraphics[height=2.8cm]{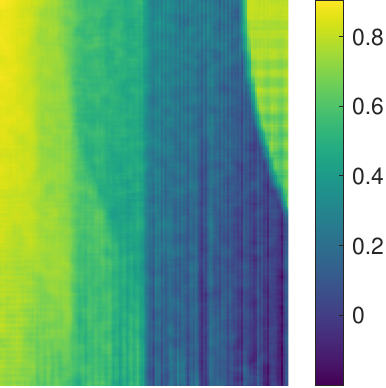}%
        \label{subfig:12MP_15_vs_12mini_BK}}
        \hfill
        \subfloat[]{\includegraphics[height=2.8cm]{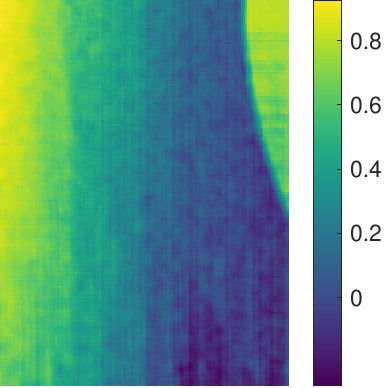}%
        \label{subfig:7MP_15_vs_12min_BK}}
        \hfill
        \subfloat[]{\includegraphics[height=2.8cm]{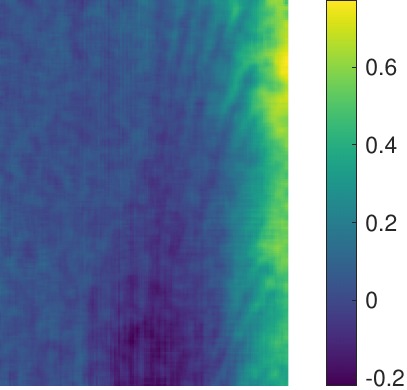}%
        \label{subfig:7_Plus_iOS11_vs_8Plus}}
        \\
        \subfloat[]{\includegraphics[height=2.8cm]{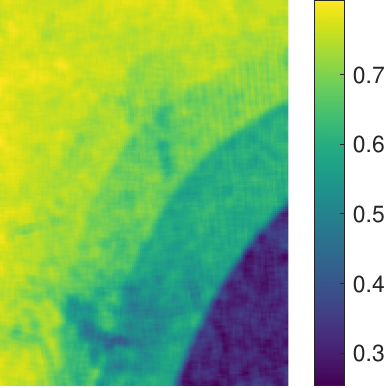}%
        \label{subfig:15_vs_11pro_JPEG}}
        \hfill
         \subfloat[]{\includegraphics[height=2.8cm]{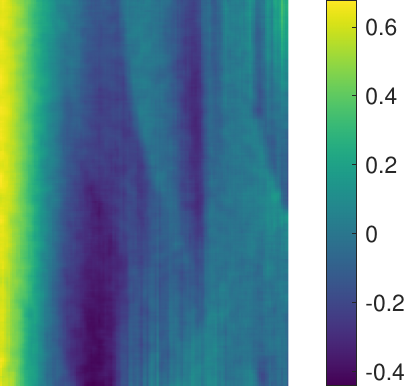}%
        \label{subfig:15_vs_16}}
        \hfill
        \subfloat[]{\includegraphics[height=2.8cm]{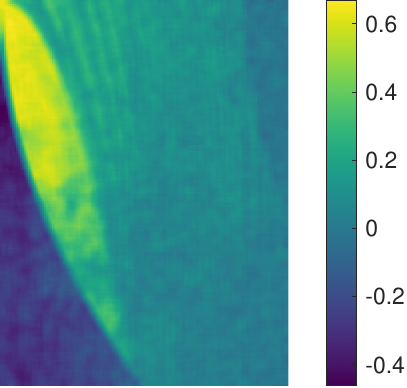}%
        \label{subfig:Photos_vs_15}}
        \caption{NCC maps illustrating partial matches between different BPs: iPhone 15 (BP \mytextcircled{6}) x iPhone 12 mini (BP \mytextcircled{5}) (a); iPhone 15 ($\hat{\mathbf{P}}_{\text{SLM}}$) x iPhone 12 mini ($\hat{\mathbf{P}}_{\text{SLM}}$) for 7MP (b); BP \mytextcircled{2} x BP \mytextcircled{3} (c); BP \mytextcircled{6} x BP \mytextcircled{4}$(\leftrightarrow)$ (d); BP \mytextcircled{6} x BP \mytextcircled{7} (e); BP \mytextcircled{6} x BP \mytextcircled{8} (f).}
        \label{fig:correlation_maps}
\end{figure}

Apple's BP in portrait images varies with resolution, iPhone model, and iOS version. While an n-depth examination of different resolutions, aspect ratios, and other BP characteristics is provided in \cite[Sect.~4.1]{SUP_MATERIAL_2025}, here we focus on how the BP changes over time for specific iPhone model/iOS combinations. As an example, although the BPs from the iPhones 15 and 12 mini share statistical properties, their patterns differ. The NCC map between their SLM-based BPs (computed as described below in Sect.~\ref{subsec:BP_PRNU_attribution}, treating one BP as a residue) shows poor correlation in the right half (Fig.~\ref{subfig:12MP_15_vs_12mini_BK}), except for a small patch in the upper-right corner. The presence of arc patterns suggests that BP generation is algorithmically driven, with both model- and software-specific variations, as further discussed below. Interestingly, like the iPhone 15 (cf. \cite[Sect.~4.1]{SUP_MATERIAL_2025}), the iPhone 12 mini exhibits no correlation between its 12MP and 7MP BPs. However, the NCC map between the BPs $\hat{\mathbf{P}}_{\text{SLM}}$  of the iPhone 15 and iPhone 12 mini at 7MP resolution (see Fig.~\ref{subfig:7MP_15_vs_12min_BK}) closely resembles that obtained at 12MP (Fig.~\ref{subfig:12MP_15_vs_12mini_BK}). This observation indicates that Apple may use a distinct BP algorithm for front (7MP) and rear (24/12MP) cameras, likely in a seed/model-dependent manner. From this point onward, we focus exclusively on 12MP portrait images, as the findings directly extend to 24MP ones (for models supporting them, i.e., starting from the iPhone 15 series) given that their BPs are spatially aligned and scaled versions of each other when captured with rear cameras. In contrast, portrait images from front cameras (7MP or higher) exhibit distinct BP characteristics and will be analyzed in future work.

Motivated by these initial findings, we analyzed 12MP portrait images from all iPhone models supporting portrait mode to track distinct iPhone BP variants. The complete list of devices (including iOS versions, number of portrait images, and data sources) is available in the technical report \cite[Tab.~6]{SUP_MATERIAL_2025}. The dataset includes 36 iPhone models (21 represented by at least two distinct devices, spanning up to the latest iPhone 17 series), 99 unique devices, multiple iOS versions (from iOS 10 to iOS 26), and 2,298 portrait images.\footnote{The dataset is available at \href{https://github.com/dvazquezpadin/apple-sdnp}{https://github.com/dvazquezpadin/apple-sdnp}} These were collected from prior datasets \cite{IULIANI_2021, ALBISANI_2021}, a locally organized contest \cite{IPREMIO_CONTEST}, Flickr \cite{FLICKR_API}, and GSMArena \cite{GSMARENA} which provided out-of-camera samples for all supported models, along with three in-house devices (iPhone 12 mini, 15, and 16) used for controlled acquisitions.\footnote{Only images whose EXIF tag \texttt{CustomRendered} was set to either \texttt{Portrait} or \texttt{Portrait HDR} were considered Apple portrait images.}

From this dataset, we identified seven distinct BPs linked to different iPhone model/iOS combinations. For clarity, these are indexed sequentially (e.g., BP~\mytextcircled{1}), as illustrated in the BP compatibility map in Tab.~\ref{tab:diff_BP_versions}. Importantly, BPs are not strictly tied to a specific model but may change with the iOS version. For example, the iPhone 7 Plus (the first model with portrait mode released in 2016) exhibits BP \mytextcircled{1} under iOS 10 and BP \mytextcircled{2} under iOS 11, with no spatial correlation between them. The iPhone 8 Plus (2017, iOS 11) introduced BP \mytextcircled{3}, which shows local correlation with BP \mytextcircled{2}, as illustrated in Fig.~\ref{subfig:7_Plus_iOS11_vs_8Plus}. This partial match is denoted by the symbol \mcmark~in Tab.~\ref{tab:diff_BP_versions}.

The iPhone X (2017) initially shared BP \mytextcircled{3} under iOS 11 but adopted BP \mytextcircled{4} with iOS 12, which later appeared in the iPhone XR/XS/X Max (2018) and iPhone 11 series (2019, iOS 13). BP \mytextcircled{4} shows no correlation with earlier BPs but partially aligns (when horizontally flipped, $(\leftrightarrow)$) with BP \mytextcircled{5}, introduced with the iPhone 12 series (2020, iOS 14). The iPhone 11 series adopted BP \mytextcircled{5} starting with iOS 16 (possibly as early as iOS 14 or 15, though no samples from those versions were available).

The iPhone 13 series (2021, iOS 15) initially used BP \mytextcircled{5} but transitioned to BP \mytextcircled{6} with the iPhone 14 series (2022, iOS 16). As shown in Fig.~\ref{subfig:12MP_15_vs_12mini_BK}, BP \mytextcircled{5} and BP \mytextcircled{6} yield a partially matching NCC map, represented by the symbol \bcmark~in Tab.~\ref{tab:diff_BP_versions}. BP \mytextcircled{6} persisted in the iPhone 15 series (2023, iOS 17, excluding Pro/Pro Max models), showing high correlation with BP \mytextcircled{4} when horizontally flipped (see Fig.~\ref{subfig:15_vs_11pro_JPEG}) a partial match denoted as \ccmark$(\leftrightarrow)$ in Tab.~\ref{tab:diff_BP_versions}. Similarly, the iPhone X and SE (2nd gen.) adopted BP \mytextcircled{6}$(\leftrightarrow)$ under iOS 16.

Starting with the iPhone 15 Pro and 15 Pro Max (2023, iOS 17), a new BP \mytextcircled{7} emerged, present in the latest iPhone 16 (2024, iOS 18) and 17 series (2025, iOS 26, including the iPhone Air edition). BP \mytextcircled{7} shows structural similarities and arc patterns with previous BPs, as revealed by the NCC map between BP \mytextcircled{7} and BP \mytextcircled{6} in Fig.~\ref{subfig:15_vs_16}; this partial match is denoted by \gcmark~in Tab.~\ref{tab:diff_BP_versions}. Finally, an eighth pattern, BP \mytextcircled{8}, was detected in certain iPhone 8 Plus and iPhone X samples, showing local correlations and arc patterns with BP \mytextcircled{6} (Fig.~\ref{subfig:Photos_vs_15}). However, since these images contain editing-related EXIF tags (e.g., \texttt{IPTC Digest}), BP \mytextcircled{8} is likely dependent on editing software (possibly originating from Apple's \textit{Photos} app rather than the camera pipeline) and is therefore excluded from Tab.~\ref{tab:diff_BP_versions}.

Tab.~\ref{tab:diff_BP_versions} summarizes the BP compatibility map at 12MP resolution across the examined iPhone models, grouped by iPhone model/iOS pair and BP index (\mytextcircled{1}–\mytextcircled{7}). As noted above, BP \mytextcircled{8} is excluded, as it is conjectured to be software-based. In this table, \xmark~indicates no match between BPs (i.e., no strong local correlation in the NCC map), whereas \cmark~denotes a complete match (i.e., all model/iOS pairs under the same BP index share the same BP). Partial matches are categorized into four classes, as described earlier: \mcmark~corresponds to an NCC map like Fig.~\ref{subfig:7_Plus_iOS11_vs_8Plus}; \bcmark~to Fig.~\ref{subfig:12MP_15_vs_12mini_BK}; \ccmark~to Fig.~\ref{subfig:15_vs_11pro_JPEG}; and \gcmark~to Fig.~\ref{subfig:15_vs_16}. The symbol $(\leftrightarrow)$ denotes cases where one BP must be horizontally flipped to reveal spatial correlations in the NCC map. The resulting BP compatibility map has strong forensic relevance, as it enables inferring both the likely iPhone model and iOS version from a portrait image of unknown origin. Notably, cases where distinct BPs correspond to the same model but different iOS versions (e.g., iPhone 7 Plus, iPhone X, or iPhone 11/13 series) can offer valuable clues for timeline reconstruction and device identification.

\begingroup
\setlength{\tabcolsep}{3.3pt} 
\renewcommand{\arraystretch}{2.5} 
\begin{table*}[t]
\caption{BP Compatibility across Different iPhone Models at 12MP Resolution.}
\label{tab:diff_BP_versions}
\centering
\begin{adjustbox}{width=\linewidth}
\begin{tabular}{clc!{\color{gray!60}\vrule width 0.4pt}c!{\color{gray!60}\vrule width 0.4pt}c!{\color{gray!60}\vrule width 0.4pt}c!{\color{gray!60}\vrule width 0.4pt}c!{\color{gray!10}\vrule width 0.2pt}c!{\color{gray!60}\vrule width 0.4pt}c!{\color{gray!10}\vrule width 0.2pt}c!{\color{gray!10}\vrule width 0.2pt}c!{\color{gray!10}\vrule width 0.2pt}c!{\color{gray!10}\vrule width 0.2pt}c!{\color{gray!10}\vrule width 0.2pt}c!{\color{gray!60}\vrule width 0.4pt}c!{\color{gray!60}\vrule width 0.4pt}c!{\color{gray!10}\vrule width 0.2pt}c!{\color{gray!10}\vrule width 0.2pt}c!{\color{gray!10}\vrule width 0.2pt}c!{\color{gray!60}\vrule width 0.4pt}c!{\color{gray!10}\vrule width 0.2pt}c!{\color{gray!10}\vrule width 0.2pt}c}\toprule
& &\textbf{BP}&\textbf{\mytextcircled{1}}& \textbf{\mytextcircled{2}}&\textbf{\mytextcircled{3}}&\multicolumn{2}{c!{\color{gray!60}\vrule width 0.4pt}}{\textbf{\mytextcircled{4}}} & \multicolumn{6}{c!{\color{gray!60}\vrule width 0.4pt}}{\textbf{\mytextcircled{5}}} & \textbf{\mytextcircled{6} $(\leftrightarrow)$} & \multicolumn{4}{c!{\color{gray!60}\vrule width 0.4pt}}{\textbf{\mytextcircled{6}}} & \multicolumn{3}{c}{\textbf{\mytextcircled{7}}}\\\cmidrule{3-22}
& &\makecell[bc]{\textbf{iPhone}\\\textbf{Model}} &7 Plus &7 Plus &\makecell[bl]{8 Plus\\ X} &\makecell[bl]{X\\XR, XS\\ XS Max} &\makecell[bl]{11, 11 Pro\\11 Pro Max\\ SE (2nd)} &12 Pro Max &\makecell[bl]{12\\ 12 Pro} &12 mini &\makecell[bl]{13, 13 mini\\13 Pro, SE (3rd)\\ 13 Pro Max} & \makecell[bl]{11\\11 Pro Max} & 11 Pro &\makecell[bl]{X\\SE (2nd)} & \makecell[bl]{13 Pro\\ 14 Plus\\ 14 Pro} &\makecell[bl]{13\\ 14\\ 14 Pro Max} &\makecell[bl]{13 mini\\ 13 Pro Max\\ 15 Plus} &15 &\makecell[bl]{15 Pro\\ 15 Pro Max} &\makecell[bl]{16, 16 Plus\\16e, 16 Pro\\ 16 Pro Max} &\makecell[bl]{17, 17 Pro\\ 17 Pro Max\\ Air} \\\cmidrule{3-22}
\textbf{BP} & \textbf{iPhone Model} &\textbf{iOS} &10 &11 &11 &12, 13 &13 &14 &14, 17 &14-17 &15 &16, 17 &17 &16 &16 &16, 17 &17 &17-26 &17 &18 &26 \\\midrule
\textbf{\mytextcircled{1}} & 7 Plus &10 &\cmark &\xmark &\xmark &\multicolumn{2}{c}{\xmark} &\multicolumn{6}{!{\color{gray!60}\vrule width 0.4pt}c!{\color{gray!60}\vrule width 0.4pt}}{\xmark} &\xmark &\multicolumn{4}{c}{\xmark} &\multicolumn{3}{!{\color{gray!60}\vrule width 0.4pt}c}{\xmark} \\\arrayrulecolor{gray!60}\hline
\textbf{\mytextcircled{2}} & 7 Plus &11 &\xmark &\cmark &\mcmark &\multicolumn{2}{c}{\xmark} &\multicolumn{6}{!{\color{gray!60}\vrule width 0.4pt}c!{\color{gray!60}\vrule width 0.4pt}}{\xmark} &\xmark &\multicolumn{4}{c}{\xmark} &\multicolumn{3}{!{\color{gray!60}\vrule width 0.4pt}c}{\xmark} \\\arrayrulecolor{gray!60}\hline
\textbf{\mytextcircled{3}} & 8 Plus, X &11 &\xmark &\mcmark &\cmark &\multicolumn{2}{c}{\xmark} &\multicolumn{6}{!{\color{gray!60}\vrule width 0.4pt}c!{\color{gray!60}\vrule width 0.4pt}}{\xmark} &\xmark &\multicolumn{4}{c}{\xmark} &\multicolumn{3}{!{\color{gray!60}\vrule width 0.4pt}c}{\xmark} \\\arrayrulecolor{gray!60}\hline
\textbf{\mytextcircled{4}} & \makecell[l]{X, XR, XS, XS Max\\\makecell[l]{11, 11 Pro, 11 Pro Max\\ SE (2nd)}} & \makecell[c]{12, 13\\\vspace{0.3cm}\multirow{2}{*}{13}} & \xmark & \xmark & \xmark &\multicolumn{2}{c}{\cmark} & \multicolumn{6}{!{\color{gray!60}\vrule width 0.4pt}c!{\color{gray!60}\vrule width 0.4pt}}{\bcmark $(\leftrightarrow)$} & \ccmark & \multicolumn{4}{c}{\ccmark $(\leftrightarrow)$} &  \multicolumn{3}{!{\color{gray!60}\vrule width 0.4pt}c}{\gcmark $(\leftrightarrow)$}\\\arrayrulecolor{gray!60}\hline
\textbf{\mytextcircled{5}} &\makecell[l]{12 Pro Max\\12, 12 Pro\\12 mini\\\makecell[l]{13, 13 mini, 13 Pro\\ 13 Pro Max, SE (3rd)}\\11, 11 Pro Max\\11 Pro} &\makecell[c]{14\\14, 17\\14-17\\\vspace{0.3cm}\multirow{2}{*}{15}\\16, 17\\17} & \xmark & \xmark & \xmark &\multicolumn{2}{c}{\bcmark $(\leftrightarrow)$} & \multicolumn{6}{!{\color{gray!60}\vrule width 0.4pt}c!{\color{gray!60}\vrule width 0.4pt}}{\cmark} &\bcmark $(\leftrightarrow)$ & \multicolumn{4}{c}{\bcmark} & \multicolumn{3}{!{\color{gray!60}\vrule width 0.4pt}c}{\gcmark}\\\arrayrulecolor{gray!60}\hline
\textbf{\mytextcircled{6} $(\leftrightarrow)$} &X, SE (2nd) & 16 & \xmark & \xmark & \xmark &\multicolumn{2}{c}{\ccmark} & \multicolumn{6}{!{\color{gray!60}\vrule width 0.4pt}c!{\color{gray!60}\vrule width 0.4pt}}{\bcmark $(\leftrightarrow)$} & \cmark & \multicolumn{4}{c}{\cmark $(\leftrightarrow)$} & \multicolumn{3}{!{\color{gray!60}\vrule width 0.4pt}c}{\gcmark $(\leftrightarrow)$}\\\arrayrulecolor{gray!60}\hline
\textbf{\mytextcircled{6}} &\makecell[l]{13 Pro, 14 Plus, 14 Pro\\13, 14, 14 Pro Max\\13 mini, 13 Pro Max, 15 Plus\\15} & \makecell[c]{16\\16, 17\\17\\17-26} & \xmark & \xmark & \xmark & \multicolumn{2}{c}{\ccmark $(\leftrightarrow)$} & \multicolumn{6}{!{\color{gray!60}\vrule width 0.4pt}c!{\color{gray!60}\vrule width 0.4pt}}{\bcmark}  & \cmark$(\leftrightarrow)$ & \multicolumn{4}{c}{\cmark} &\multicolumn{3}{!{\color{gray!60}\vrule width 0.4pt}c}{\gcmark} \\\arrayrulecolor{gray!60}\hline
\textbf{\mytextcircled{7}} &\makecell[l]{15 Pro, 15 Pro Max\\16, 16 Plus, 16 Pro\\ 16 Pro Max, 16e\\17, 17 Pro, 17 Pro Max, Air} & \makecell[c]{17\\\vspace{0.3cm}\multirow{2}{*}{18}\\26} & \xmark & \xmark & \xmark & \multicolumn{2}{c}{\gcmark $(\leftrightarrow)$} & \multicolumn{6}{!{\color{gray!60}\vrule width 0.4pt}c!{\color{gray!60}\vrule width 0.4pt}}{\gcmark} & \gcmark$(\leftrightarrow)$ &\multicolumn{4}{c}{\gcmark} &\multicolumn{3}{!{\color{gray!60}\vrule width 0.4pt}c}{\cmark} \\
\arrayrulecolor{black}\bottomrule
\end{tabular}
\end{adjustbox}
\end{table*}
\endgroup

\section{BP-based Forensic Applications}
\label{sec:BP_applications}

In this section, we propose BP-enabled forensic applications, including BP detection and identification (Sect.~\ref{subsec:BP_detection_identification}), a BP-aware approach for PRNU-based camera source verification (Sect.~\ref{subsec:BP_PRNU_attribution}), and, finally, a proof-of-concept for forgery localization using BP-aware PRNU analysis (Sect.~\ref{subsec:BP_PRNU_forgery_localization}).

\subsection{Apple's BP Detection and Identification}
\label{subsec:BP_detection_identification}

Detecting Apple's BP in an image is valuable for labeling or isolating images generated by computational imaging (e.g., portrait mode, as proposed in \cite{ALBISANI_2021, MCCLOSKEY_2022}) and for curating machine learning datasets. Identifying the specific BP also enables tracing the iPhone model and iOS version, aiding forensic investigations.

A simple detection method computes the noise residue $\mathbf{W}$ from the image under analysis via \eqref{eq:W_l} with $K=5$ (higher $K$ only benefits uniform regions), followed by the NCC in \eqref{eq:NCC} between $\mathbf{W}$ and a given BP estimate $\hat{\mathbf{P}}$. If image orientation is unknown, all $90^\circ$ rotations must be tested. Images sharing the same (spatially aligned) BP should yield a high NCC; thus, those with $\rho(\mathbf{W},\hat{\mathbf{P}})>\beta$, where $\beta$ is a predefined threshold, are flagged as Apple portrait images.

Given multiple BP versions (see Sect.~\ref{sec:BP_variations}) and potentially missing EXIF metadata, the NCC must be computed across all BP estimates and $90^\circ$ rotation variants. The BP yielding the highest NCC above $\beta$ is selected; if none exceed $\beta$, no BP match is detected. The predefined threshold $\beta=0.0072$, used in the experimental validation of Sect.~\ref{sec:exp_results}, was empirically determined to achieve a target false positive rate of $10^{-3}$ on a controlled dataset, as detailed in the technical report \cite[Sect.~5.1]{SUP_MATERIAL_2025}.

\begin{figure}[t]
        \centering
        \subfloat[ISO 400]{\includegraphics[height=2.32cm]{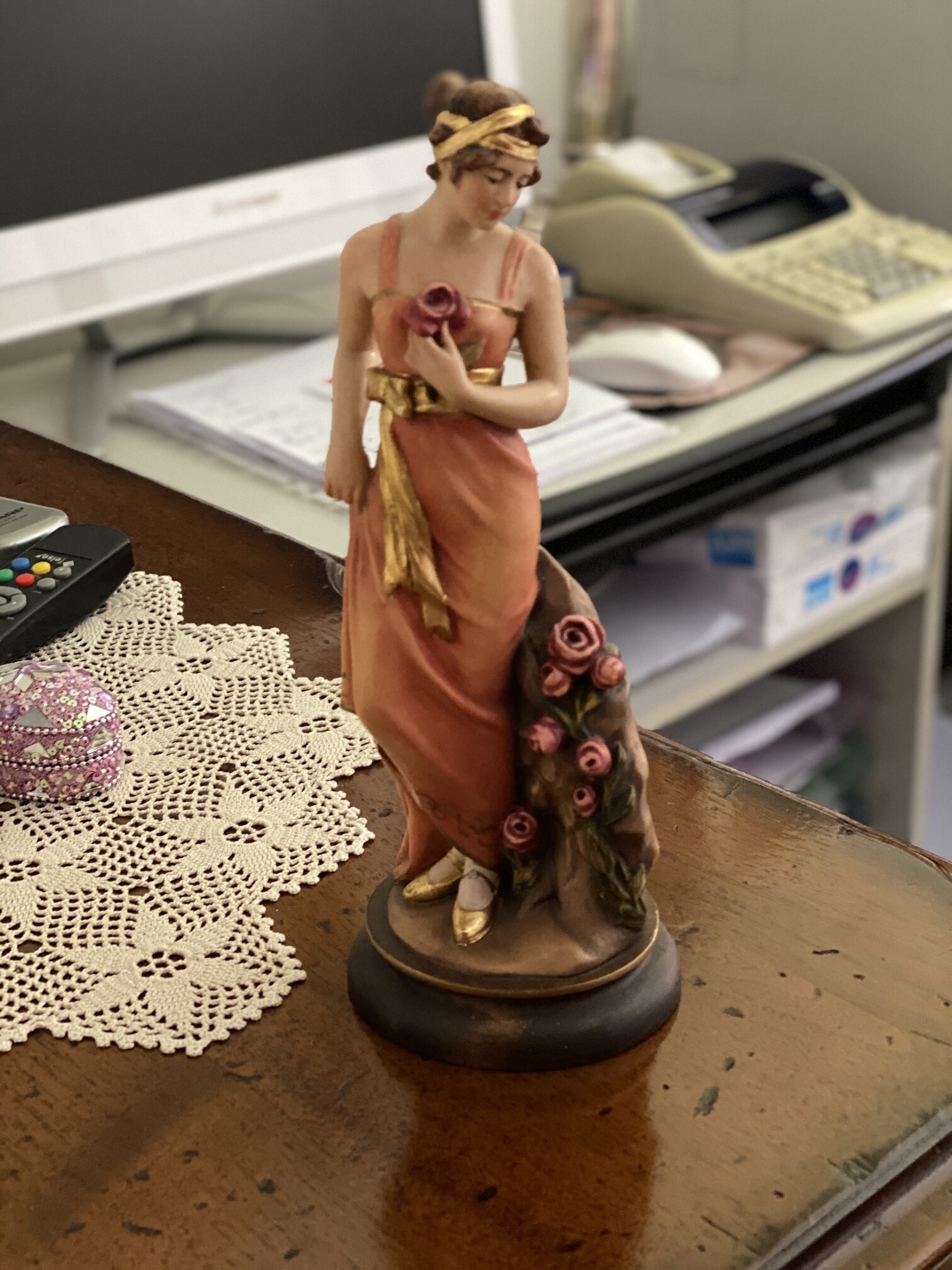}%
        \hspace{0.01pt}
        \includegraphics[height=2.32cm]{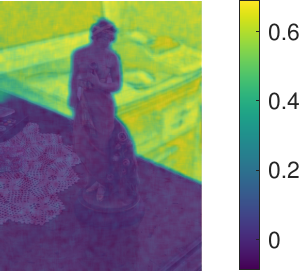}%
        \label{subfig:01_albisani_example}}
        \hfill\hfill
        \subfloat[ISO 125]{\includegraphics[height=2.32cm]{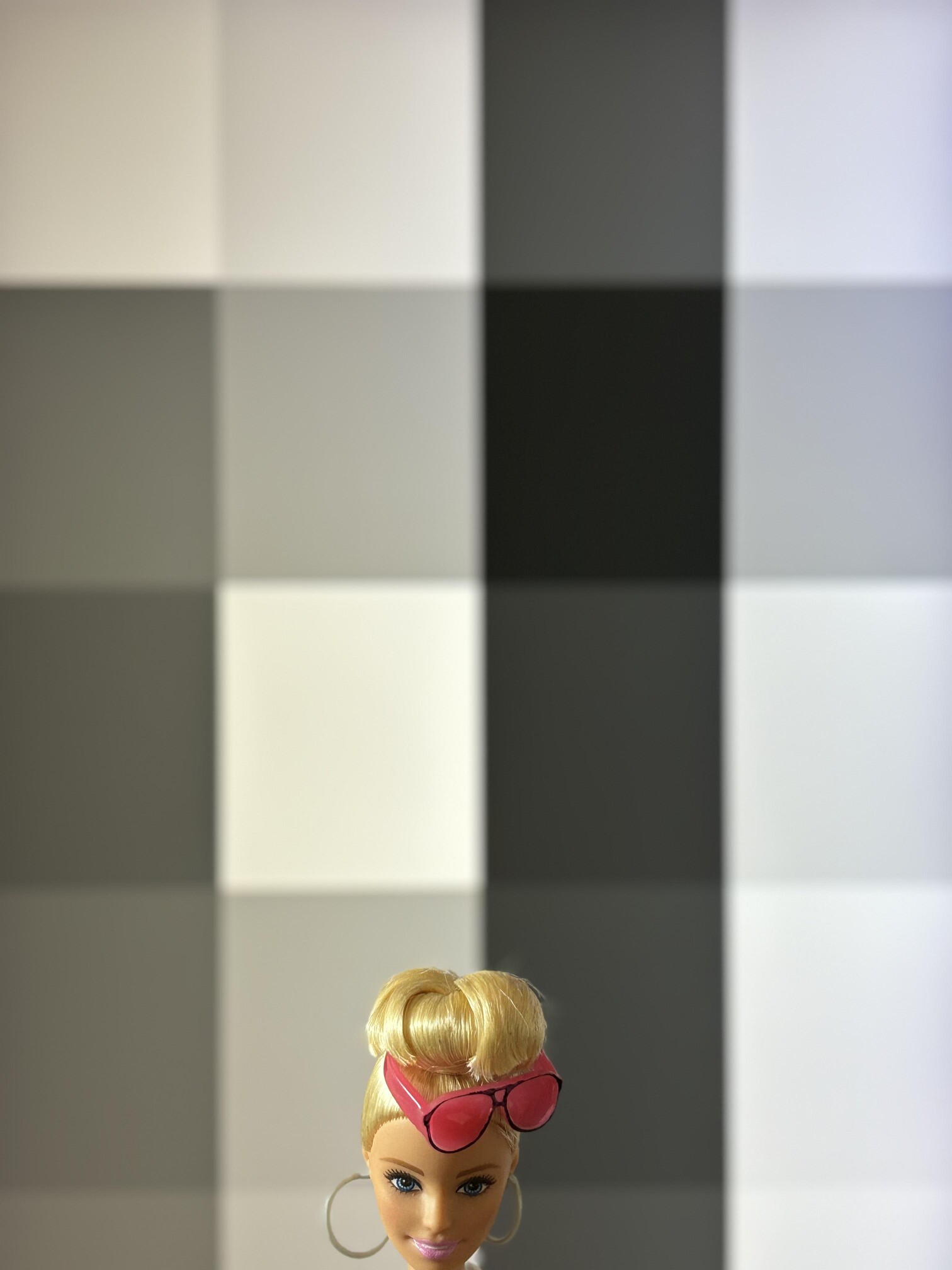}%
        \hspace{0.01pt}
        \includegraphics[height=2.32cm]{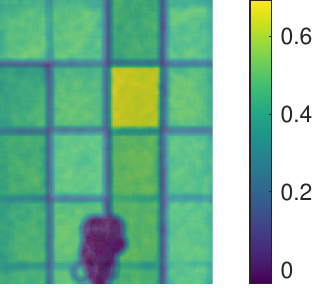}%
        \label{subfig:02_albisani_example}}
        \caption{Portrait images and their relative BP-driven NCC maps: iPhone 11 Pro Max (iOS 13.4) \cite[C22]{ALBISANI_2021} (a) and iPhone 15 (iOS 17.5.1) (b). In both cases, $\hat{\mathbf{P}}_{\text{SLM}}$ from the iPhone 15, i.e., BP~\mytextcircled{6}, is used, applying a horizontal flip, i.e., BP~\mytextcircled{6}$(\leftrightarrow)$, in (a).}
        \label{fig:BP_corr_map}
\end{figure}

\subsection{BP-aware PRNU-based Camera Source Verification}
\label{subsec:BP_PRNU_attribution}

To address the bias that Apple's SDNP induces in PRNU-based source verification (cf. Fig.~\ref{fig:baseline_results_FORLAB} in Sect.~\ref{subsec:PRNU_matching}), we exploit the fact that regions containing the SDNP can be localized using a matching BP estimate. In particular, a BP-driven NCC map is computed between a given image residue $\mathbf{W}$ (obtained through \eqref{eq:W_l}) and a BP estimate $\hat{\mathbf{P}}$, with local NCC values evaluated over $21 \times 21$ blocks using \eqref{eq:NCC}. For blocks near image boundaries, missing samples are padded by mirror reflection. The resulting map is then smoothed with a box filter ($K=5$) and resized via nearest-neighbor interpolation to match the dimensions of $\mathbf{W}$, yielding the final BP-driven NCC map $\mathbf{R}$. For illustration, we used the BP estimate from 12MP SLM portrait images of the iPhone 15, $\hat{\mathbf{P}}_{\text{SLM}}$, to generate the NCC map $\mathbf{R}$ for two portrait images: one from device C22 in the dataset from \cite{ALBISANI_2021} (where $\hat{\mathbf{P}}_{\text{SLM}}^{(\leftrightarrow)}$ is used) and another from the same iPhone 15 device. The overlaid maps (Figs.~\ref{subfig:01_albisani_example} and \ref{subfig:02_albisani_example}) reveal regions where the SDNP is present, as indicated by higher NCC values.

Leveraging this map, we propose filtering out SDNP-affected regions during PRNU extraction and detection. When a matching BP estimate is available, it is used directly; otherwise, the method in Sect.~\ref{subsec:BP_detection_identification} is used to identify the best-matching BP (even if the highest NCC value falls below the threshold $\beta$). From this, we compute the BP-driven NCC map $\mathbf{R}$ between the image residue $\mathbf{W}$ and the BP estimate $\hat{\mathbf{P}}$, applying a predefined threshold $\alpha$ to create a binary mask $\mathbf{M}^{\text{(PRNU)}}$. This mask excludes regions where the BP is likely present by setting $M_{i,j}^{\text{(PRNU)}} = 1$ where $R_{i,j} \leq \alpha$, and $M_{i,j}^{\text{(PRNU)}} = 0$ otherwise. As reported in the technical report \cite[Sect.~5.2]{SUP_MATERIAL_2025}, the predefined threshold was empirically set to $\alpha=0.07$, corresponding to an equal error rate of $0.13$ for the selected $21\times21$ block size, and is used in the experimental validation of Sect.~\ref{sec:exp_results}. Applying the resulting mask $\mathbf{M}^{\text{(PRNU)}}$ in \eqref{eq:hat_K} yields a BP-aware PRNU estimate:
\begin{equation}
        \hat{\mathbf{K}}^\prime=\left(\sum_{l=1}^{L}\mathbf{W}_l^\prime\circ\mathbf{Z}_l^\prime\right)\circ\left(\sum_{l=1}^{L}\mathbf{Z}_l^\prime\circ\mathbf{Z}_l^\prime\right)^{\circ-1},
        \label{eq:hat_K_prime}
\end{equation}
where $\mathbf{W}_l^\prime=\mathbf{M}_l^{\text{(PRNU)}}\circ\mathbf{W}_l$ and $\mathbf{Z}_l^\prime=\mathbf{M}_l^{\text{(PRNU)}}\circ\mathbf{Z}_l$ represent the $l$th masked residue and masked image, respectively, excluding regions where the BP under analysis is detected. Note the slight abuse of notation here: in this context, $\mathbf{Z}_l$ can represent either a portrait or a non-portrait image. If non-portrait images are processed, we expect $\mathbf{M}_l^{\text{(PRNU)}}$ to be a matrix entirely composed of ones. Given the BP-aware PRNU estimate $\hat{\mathbf{K}}^\prime$, we modify the similarity measure defined in \eqref{eq:eta} to incorporate the SDNP masking. The test for checking the PRNU presence in a test image $\mathbf{Z}_t$ becomes:
\begin{equation}
        \eta^\prime\triangleq N\cdot \text{ssq}\left(\rho\left(\mathbf{W}_t^\prime,\hat{\mathbf{K}}^\prime\circ\mathbf{Z}_t^\prime\right)\right)\underset{\mathcal{H}_0}{\overset{\mathcal{H}_1}{\gtrless}}\tau^\prime,
        \label{eq:eta_prime}
\end{equation}
where $\mathbf{W}_t^\prime=\mathbf{W}_t\circ\mathbf{M}_t^{\text{(PRNU)}}$ and $\mathbf{Z}_t^\prime=\mathbf{Z}_t\circ\mathbf{M}_t^{\text{(PRNU)}}$. Similarly to BP alignment, when the image orientation is unknown, we evaluate all possible $90^\circ$ rotations of $\hat{\mathbf{K}}^\prime$ and select the highest resulting $\eta^\prime$ value to ensure proper alignment with the PRNU.

\subsection{Forgery Localization Using BP-aware PRNU Analysis}
\label{subsec:BP_PRNU_forgery_localization}

The BP-driven NCC map introduced in the previous section can be used not only to identify and exclude bokeh-affected regions, as in the PRNU matching use case, but also to detect areas lacking the expected BP correlation, for example due to object removal in a bokeh-affected region. Building on this idea, a proof-of-concept for forgery localization using a combined BP–PRNU analysis is presented in our technical report \cite[Sect.~5.4]{SUP_MATERIAL_2025}. In this approach, BP-driven NCC maps reveal regions lacking the expected BP correlation; however, such absence alone does not distinguish between intentional omission by Apple's algorithm and image tampering. To mitigate this ambiguity, we also compute PRNU-driven NCC maps from the capturing device. Although the PRNU signal is weaker than the BP and produces maps with less spatial detail, it provides complementary information by indicating which regions still retain the device's PRNU and thus are unlikely to be manipulated. By normalizing and combining both maps, regions with low correlation values can be visualized as a heatmap, highlighting potential forgeries. Some false positives may arise due to noise in the PRNU map (see \cite[Fig.~17]{SUP_MATERIAL_2025}), particularly in images subjected to heavy post-processing such as recompression or global editing. Nonetheless, this proof-of-concept demonstrates that forgery localization is feasible in real-world conditions whenever the manipulated region originally contained Apple's BP and a reliable device-specific PRNU estimate is available. Future work will explore improved PRNU estimation and noise reduction strategies to further enhance localization accuracy.

\section{Experimental Results}
\label{sec:exp_results}

We conducted several experiments to validate our BP-enabled forensic applications, covering BP detection and identification, PRNU collision mitigation, comparison with Baracchi \emph{et al.} \cite{BARACCHI_2021}, and robustness against post-processing.\footnote{The code is available at \href{https://github.com/dvazquezpadin/apple-sdnp}{https://github.com/dvazquezpadin/apple-sdnp}}

\subsection{Performance of Apple's BP Detection and Identification}
\label{subsec:performance_BP_det_and_id}

To validate our method for Apple's BP detection and identification (Sect.~\ref{subsec:BP_detection_identification}), after setting the predefined threshold to $\beta=0.0072$ (whose computation is detailed in~\cite[Sect.~5.1]{SUP_MATERIAL_2025}), we conducted tests in a controlled scenario using a balanced dataset of 4,596 images at 12MP resolution. The negative set consists of 2,298 non-portrait images from \cite{IULIANI_2021} (cf. \cite[Tab.~5]{SUP_MATERIAL_2025}), including samples from multiple iPhone and Samsung models. The positive set, of equal size, comprises 2,298 Apple portrait images (cf. \cite[Tab.~6]{SUP_MATERIAL_2025}) covering, to the best of our knowledge, all 36 iPhone models supporting portrait mode (from the iPhone 7 Plus to the latest iPhone 17 Series, including the Air variant). For 21 of these models, images were available from two distinct devices (with 15 models represented by more than two devices, and up to 7 or 9 in some cases) across a range of iOS versions (from iOS 10 to iOS 26), enabling an evaluation of the method’s robustness to iOS-dependent BP variations. For this experiment, we used all available BP estimates (i.e., the eight detailed in Sect.~\ref{sec:BP_variations}). In all cases, NL-based BPs $\hat{\mathbf{P}}_{\text{NL}}$ were employed, except for BPs \mytextcircled{5} and \mytextcircled{7}, where SLM-based BPs $\hat{\mathbf{P}}_{\text{SLM}}$ were used, extracted both at ISO 125 from the iPhone 12 mini (D51) and iPhone 16 (D88), respectively.

Under this setup, our detector correctly classified all non-Apple portrait images except one, corresponding to an iPhone 8 photo edited with Photoshop according to its metadata, resulting in a False Positive Rate (FPR) of $4.3\cdot10^{-4}$. Among the positive samples, the detector achieved a high True Positive Rate (TPR) of $0.9761$. The few missed detections correspond to images where the bokeh region is extremely small (or seemingly absent despite being labeled as portrait images) thus providing insufficient SDNP evidence. Regarding BP identification, the method correctly matched the corresponding BP for all detected portrait images, except for a few cases involving iPhone XR, XS, and XS Max devices, which intentionally modify the simulated aperture \cite{GSMARENA-FStops}, producing a software-based variant of BP~\mytextcircled{4} with an arc pattern similar to that shown in Fig.~\ref{subfig:15_vs_11pro_JPEG} when using BP~\mytextcircled{5}.

Our approach was also evaluated on the FFHQ (in-the-wild-images) dataset \cite{KARRAS_2019} to assess its performance in an open-set, uncontrolled scenario. The analysis focused on 12MP images (427 samples in total), of which 376 could be labeled because their original Flickr uploads were still available, allowing metadata recovery (73 positive portrait samples and 303 negatives). The proposed BP detector achieved a TPR of $0.9863$, with only one missed detection (an image showing traces of editing with the \textit{VSCO} app) and one incorrect attribution (an image edited with Apple's \textit{Photos} application, which matched BP \mytextcircled{8} and exhibited characteristic arc patterns as shown in \cite[Fig.~18]{SUP_MATERIAL_2025}). Six negative samples were incorrectly classified as positives, resulting in an FPR of $1.98 \cdot 10^{-2}$. Interestingly, five of these six false positives contained a pattern associated with Apple's discontinued \textit{Aperture} software that correlated with BP \mytextcircled{3}. These findings further suggest the presence of BP-like artifacts originating from Apple's software ecosystem, which will be explored in future work. The remaining false positive corresponded to an iPhone 7 image with heavy white saturation, likely post-processed but without signs of the bokeh effect. Overall, 90 portrait images (72 labeled and 18 unlabeled) were positively identified, accounting for $21.08\%$ of the 12MP FFHQ subset, confirming that Apple's BP is clearly observable in real-world imagery. A comprehensive analysis of these results is provided in the technical report \cite[Sect.~5.5]{SUP_MATERIAL_2025}.

\subsection{Tackling PRNU Collisions with BP-aware PRNU Matching}
\label{subsec:Iuliani_2021_comparison}

\begin{figure}[t]
        \centering
        \includegraphics[width=\linewidth]{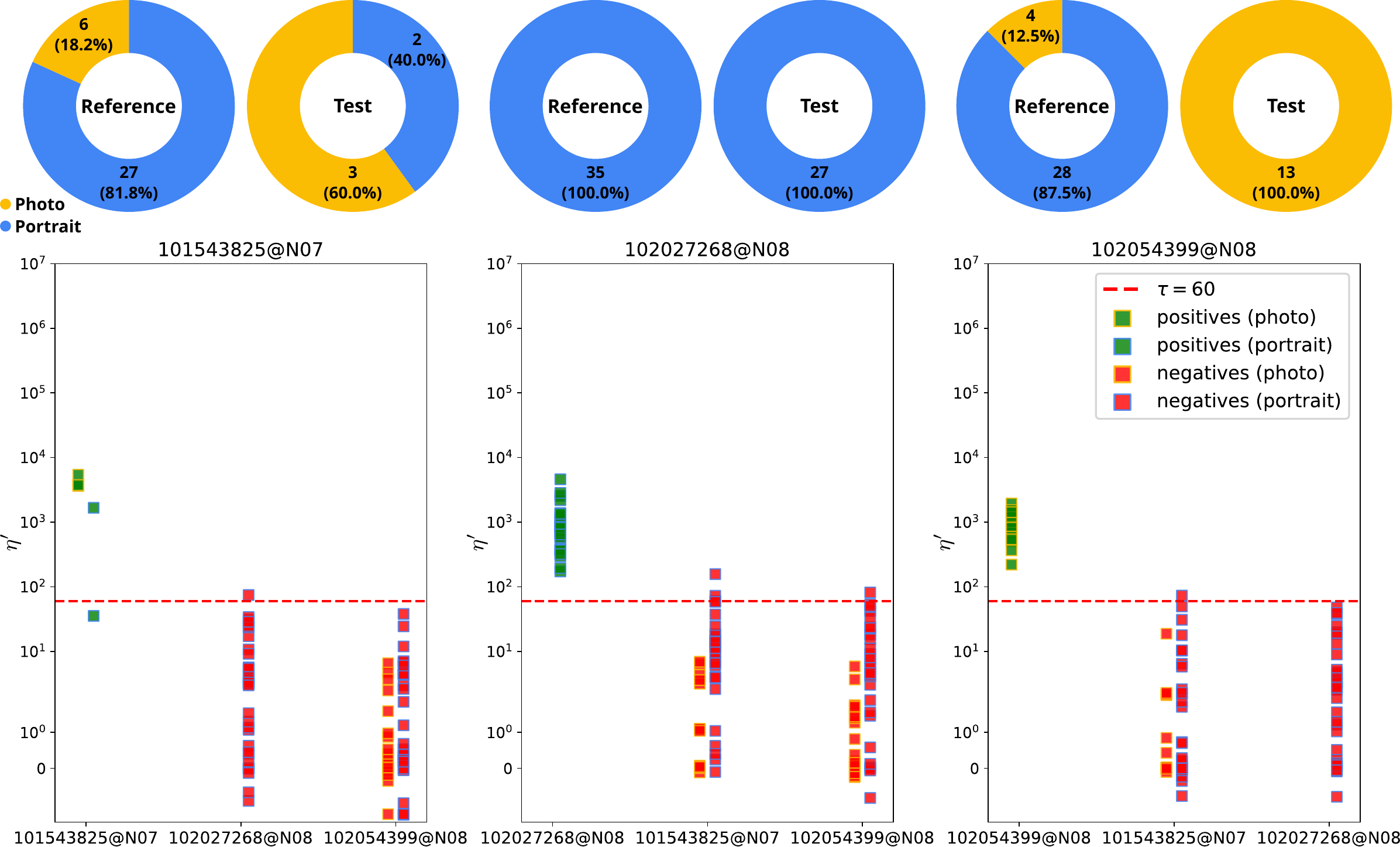}
        \caption{Results achieved using our BP-aware PRNU matching approach on the dataset employed in \cite{IULIANI_2021}. Note the sharp decrease in $\eta^\prime$ values for negative portrait images (red squares with blue borders) compared to the $\eta$ scores in Fig.~\ref{fig:baseline_results_FORLAB}, effectively eliminating most false positives at the same threshold.}
        \label{fig:exp_results_discarding_SDNP}
\end{figure}

Using the BP-aware PRNU-based camera source verification approach from Sect.~\ref{subsec:BP_PRNU_attribution} (with $K=5$ and $\alpha=0.07)$, we repeat the experiments from Sect.~\ref{subsec:PRNU_matching}, excluding 5 misaligned or different-sensor images from user \texttt{101543825@N07} (see \cite[Sect.~1]{SUP_MATERIAL_2025}). The updated results in Fig.~\ref{fig:exp_results_discarding_SDNP} show a significant reduction in the similarity measure $\eta^\prime$ for portrait-mode images compared to the original $\eta$ (Fig.~\ref{fig:baseline_results_FORLAB}). Using the original threshold ($\tau^\prime = 60$) results in few classification errors: \texttt{101543825@N07} ($\text{TPR} = 0.8$, $\text{FPR} = 0.0125$); \texttt{102027268@N08} ($\text{TPR} = 1$, $\text{FPR} = 0.0465$); and \texttt{102054399@N08} ($\text{TPR} = 1$, $\text{FPR} = 0.0132$). However, increasing the threshold to $\tau^\prime = 160$ eliminates all false positives ($\text{FPR} = 0$ for all three users), leaving only one missed detection for user \texttt{101543825@N07} ($\text{TPR} = 0.8$). These findings confirm that properly accounting for Apple's SDNP enables reliable PRNU-based source camera verification, regardless of portrait mode.

It is important to note that further adjusting the threshold $\alpha$, which defines the regions where the SDNP is present, could improve both these and subsequent results. We believe that developing a strategy to determine the optimal threshold $\alpha$ for each image would significantly enhance the overall performance of our approach, but we leave the development of this strategy for future work.

\subsection{Comparative Results of BP-aware PRNU Matching}
\label{subsec:baracchi_comparison}

Baracchi \emph{et al.} in \cite{BARACCHI_2021} were the first to address NUAs in Apple portrait images in a PRNU-based camera source verification context, showing that such NUAs correlated more strongly than the PRNU. After our analysis, however, we understand that the underlying BP of Apple's SDNP is not strictly an artifact but rather a synthetic pattern intentionally embedded by Apple. Baracchi \emph{et al.} proposed two distinct techniques, the ``weighted'' and ``binary'' methods, which use depth map information to either weigh or exclude regions in the PRNU estimate $\hat{\mathbf{K}}$ (computed using the baseline PRNU method from Sect.~\ref{subsec:PRNU_matching}), resulting in $\hat{\mathbf{K}}_{\mathbf{w}}$ and $\hat{\mathbf{K}}_{\mathbf{b}}$, respectively. For both methods, we employ the same test statistic as in \eqref{eq:eta}, substituting $\hat{\mathbf{K}}$ by $\hat{\mathbf{K}}_{\mathbf{w}}$ and $\hat{\mathbf{K}}_{\mathbf{b}}$ accordingly. In our approach (detailed in Sect.~\ref{subsec:BP_PRNU_attribution}), the image residue $\mathbf{W}$ and the binary mask $\mathbf{M}^{\text{(PRNU)}}$ are computed using $K=5$ and $\alpha=0.07$, respectively, applying the test statistic from \eqref{eq:eta_prime}. In the following, we compare Baracchi \emph{et al.}'s methods and the baseline PRNU approach with our BP-aware solution by conducting three experiments, each evaluating different BPs. 

\subsubsection{Comparative Analysis on the Dataset from \texorpdfstring{\cite{IULIANI_2021}}{}}
\label{subsubsec:comp_analysis_iuliani}

The tests are conducted on images from the 3 iPhone 11 Pro users identified with PRNU collisions in \cite{IULIANI_2021}. PRNU extraction uses the 35 images of the Reference set from user \texttt{102027268@N08}. The positive class consists of 27 test images from the same user, while the negative class includes 86 images from the other two users: \texttt{101543825@N07} (41 images, excluding 2 taken with digital zoom) and \texttt{102054399@N08} (45 images). Our approach employs the BP estimate $\hat{\mathbf{P}}_{\text{NL}}$ extracted from 40 portrait images taken by devices C19, C20, C21, and C22 in \cite{ALBISANI_2021}, whose BP~\mytextcircled{4} matches that of the iPhone 11 Pro (iOS~13) in \cite{IULIANI_2021}.

\begin{figure*}[t]
        \centering
        \subfloat[]{\begin{minipage}[t]{0.23\linewidth}\includegraphics[width=\linewidth]{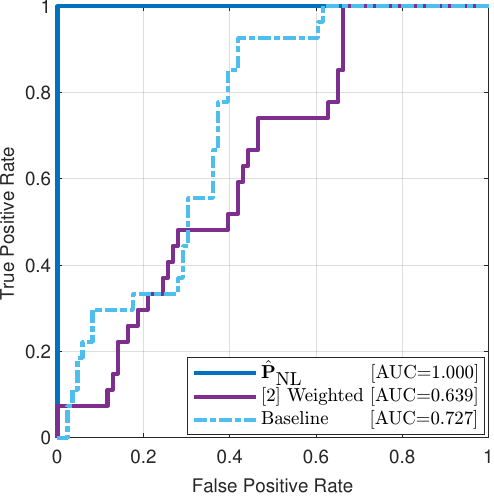}%
        \label{subfig:ROC_iphone11pro}\end{minipage}}\hfill
        \subfloat[]{\begin{minipage}[t]{0.23\linewidth}\includegraphics[width=\linewidth]{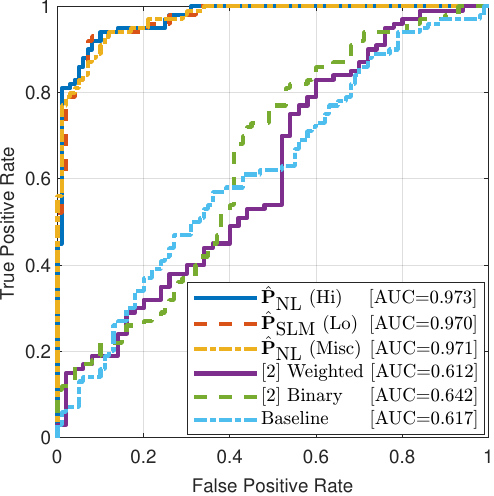}%
        \label{subfig:ROC_iphone15}\end{minipage}}\hfill
        \subfloat[]{\begin{minipage}[t]{0.23\linewidth}\includegraphics[width=\linewidth]{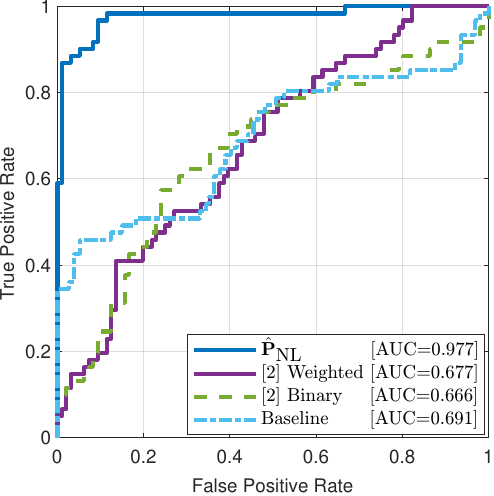}%
        \label{subfig:ROC_iphone12mini}\end{minipage}}\hfill
        \subfloat[]{\begin{minipage}[t]{0.21\linewidth}
        \vspace{-4.2cm}
        \begin{adjustbox}{width=\linewidth}
        \begin{tabular}{rccccc}\toprule
        &\multicolumn{2}{c}{\textbf{Sect.~\ref{subsubsec:comp_analysis_iuliani}}}& \multicolumn{2}{c}{\textbf{Sect.~\ref{subsubsec:comp_analysis_iphone12mini_multiuser}}} \\\cmidrule{2-5}
        &FPR &TPR &FPR &TPR \\\midrule
        $\hat{\mathbf{P}}_{\text{NL}}$&\cellcolorval{0.06} &1 &\cellcolorval{0.01} &0.80\\
        \textbf{Weighted} \cite{BARACCHI_2021} &\cellcolorval{0.66} &1 &\cellcolorval{0.75} &0.92 \\
        \textbf{Binary} \cite{BARACCHI_2021} &\textemdash &\textemdash &\cellcolorval{0.59} &0.79 \\
        \textbf{Baseline} &\cellcolorval{0.66} &1 & \cellcolorval{1} &1 \\
        \midrule
        &\multicolumn{4}{c}{\textbf{Sect.~\ref{subsubsec:BP_quality_detection_performance}}}\\\cmidrule{2-5}
        &\multicolumn{2}{c}{FPR} & \multicolumn{2}{c}{TPR}\\
        \midrule
        $\hat{\mathbf{P}}_{\text{NL}}$  (Hi) &\multicolumn{2}{c}{\cellcolorval{0.07}} & \multicolumn{2}{c}{0.89}\\
        $\hat{\mathbf{P}}_{\text{NL}}$  (Lo) &\multicolumn{2}{c}{\cellcolorval{0.09}} & \multicolumn{2}{c}{0.93}\\
        $\hat{\mathbf{P}}_{\text{NL}}$  (Misc) &\multicolumn{2}{c}{\cellcolorval{0.09}} & \multicolumn{2}{c}{0.88}\\
        \textbf{Weighted} \cite{BARACCHI_2021} &\multicolumn{2}{c}{\cellcolorval{0.98}} & \multicolumn{2}{c}{1}\\
        \textbf{Binary} \cite{BARACCHI_2021} &\multicolumn{2}{c}{\cellcolorval{0.93}} & \multicolumn{2}{c}{0.99}\\
        \textbf{Baseline} &\multicolumn{2}{c}{\cellcolorval{1}} & \multicolumn{2}{c}{1}\\
        \bottomrule
        \end{tabular}
        \end{adjustbox}\label{tab:TPR_FPR_values}\end{minipage}}
        \caption{ROC curves comparing PRNU-based source camera verification performance across different methods: baseline PRNU, the two approaches from \cite{BARACCHI_2021}, and our BP-aware solution. The plots correspond to Sect.~\ref{subsubsec:comp_analysis_iuliani} (a), Sect.~\ref{subsubsec:BP_quality_detection_performance} (b), Sect.~\ref{subsubsec:comp_analysis_iphone12mini_multiuser} (c). Panel (d) shows a table summarizing the corresponding TPR and FPR values obtained using a threshold of $\tau^\prime = 60$.}
        \label{fig:ROCs_baracchi}
\end{figure*}

The ROC curves obtained for each detector are shown in Fig.~\ref{subfig:ROC_iphone11pro}. Note that the binary method from \cite{BARACCHI_2021} is absent, because the exclusion of regions from the images captured by user \texttt{102027268@N08} results in a PRNU estimate $\hat{\mathbf{K}}_{\mathbf{b}}$ consisting entirely of zeros. Interestingly enough, the weighted method performs worse than the baseline approach (as also observed during our attempt to replicate the original results in \cite[Sect.~5.3]{SUP_MATERIAL_2025}), suggesting that its effectiveness is highly dependent on the specific scene captured and the corresponding depth map. In contrast, our approach, which focuses on localizing the presence of the BP, proves to be very effective, achieving perfect detection, as previously highlighted in Fig.~\ref{fig:exp_results_discarding_SDNP}. We believe that the primary reason for the lower performance of Baracchi \emph{et al.}'s method is its exclusive reliance on depth map information. As discussed in Sect.~\ref{subsubsec:blur_rendering}, the exact algorithm Apple uses to determine which regions will have the SDNP added remains unknown and may have evolved over time. In fact, all images in our comparative analysis were captured with newer iOS versions (i.e., iOS 13.3.1 or later) than those analyzed by Baracchi \emph{et al.} (i.e., iOS 12.1.4), which may explain why we are unable to replicate their results in \cite{BARACCHI_2021}. 

\begin{figure}[t]
        \centering
        \subfloat[]{\includegraphics[height=1.45cm]{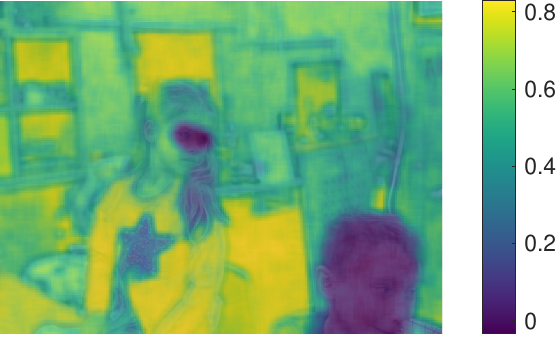}%
        \label{subfig:example_misdetection_user2}}
        \hfill\hfill
        \subfloat[]{\includegraphics[height=1.45cm]{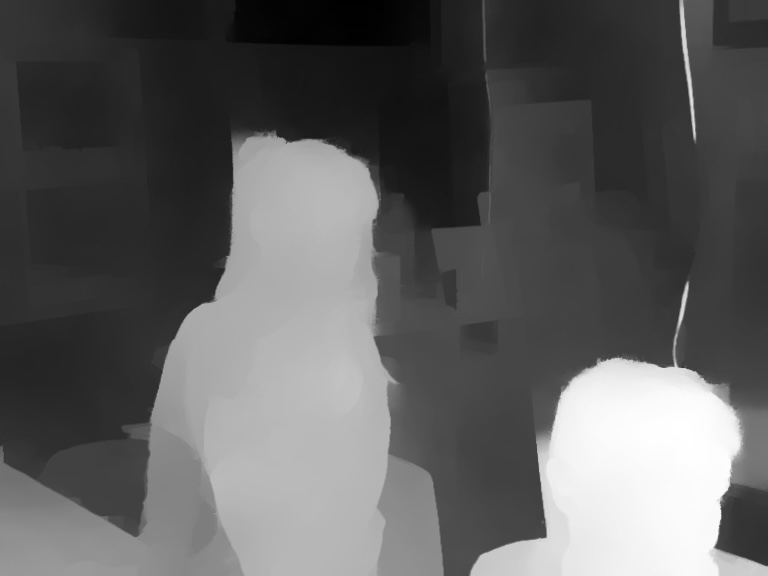}%
        \label{subfig:depth_map}}
        \hfill\hfill
        \subfloat[]{\includegraphics[height=1.45cm]{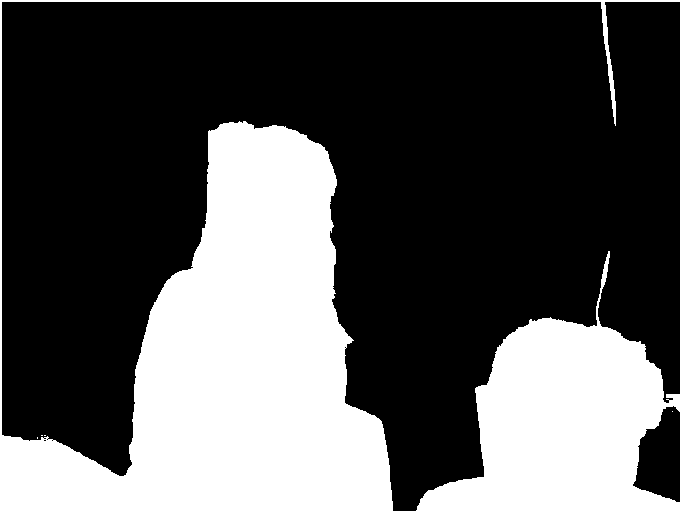}%
        \label{subfig:baracchi_mask}}
        \hfill\hfill
        \subfloat[]{\includegraphics[height=1.45cm]{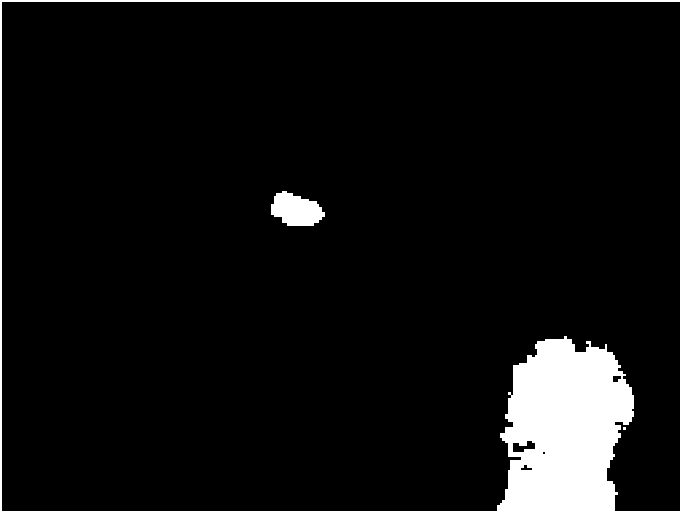}%
        \label{subfig:SDNP_mask}}
        \\\vspace{0.2cm}
        \subfloat[]{\includegraphics[height=1.45cm]{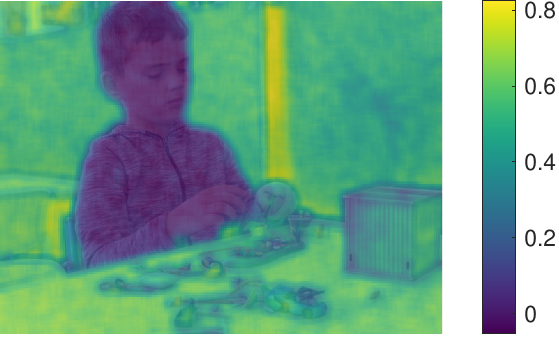}%
        \label{subfig:example_false_positive}}
        \hfill\hfill
        \subfloat[]{\includegraphics[height=1.45cm]{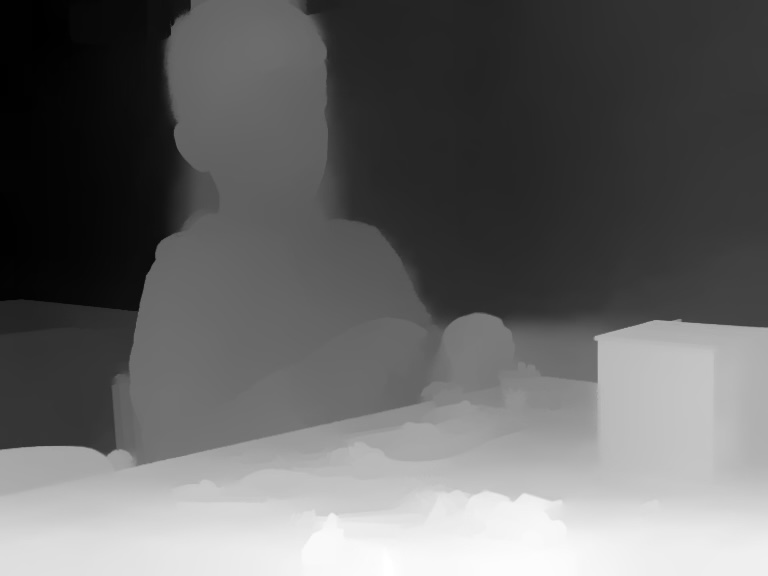}%
        \label{subfig:example_FP_depth_map}}
        \hfill\hfill
        \subfloat[]{\includegraphics[height=1.45cm]{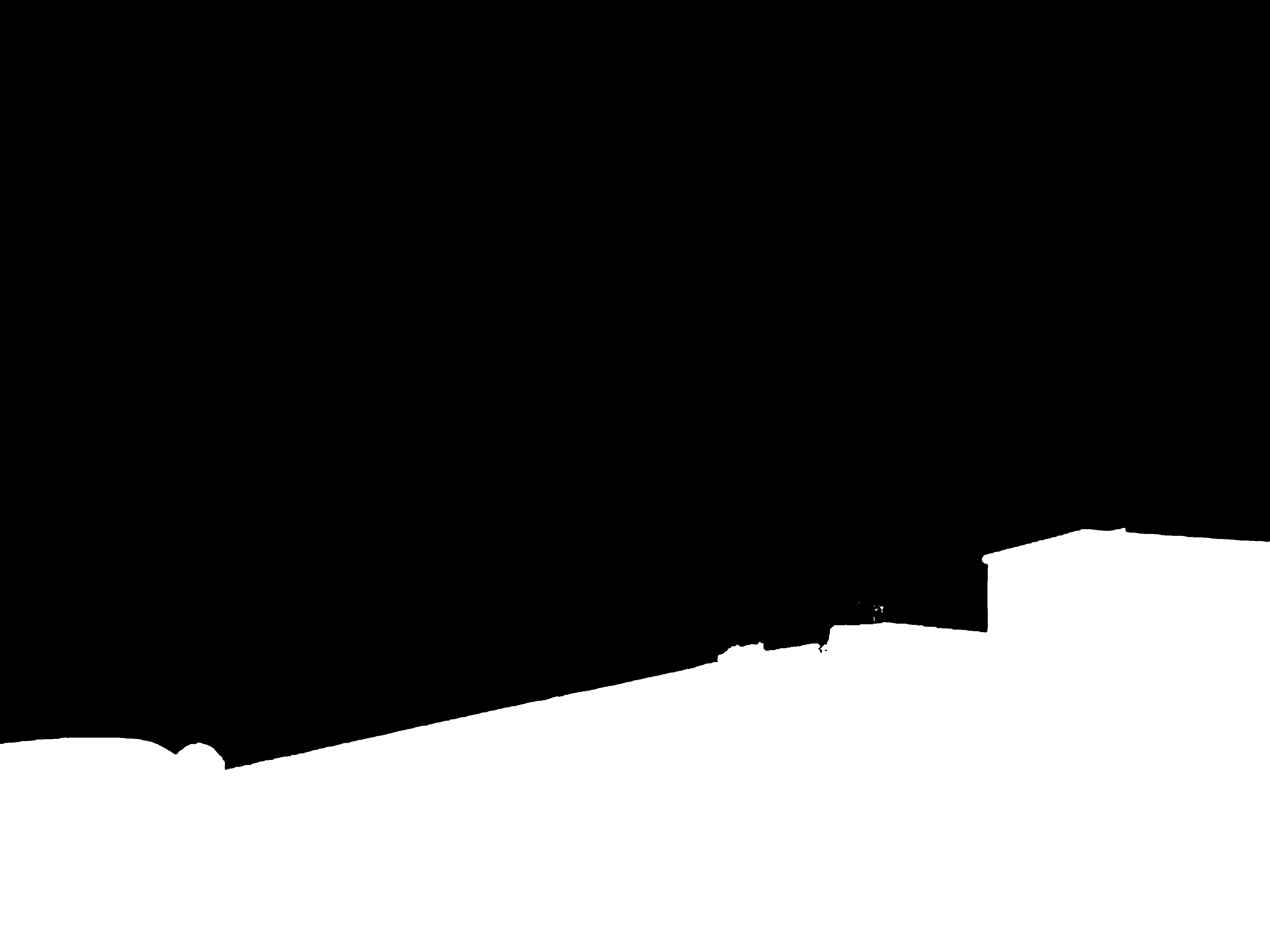}%
        \label{subfig:example_FP_binary_mask}}
        \hfill\hfill
        \subfloat[]{\includegraphics[height=1.45cm]{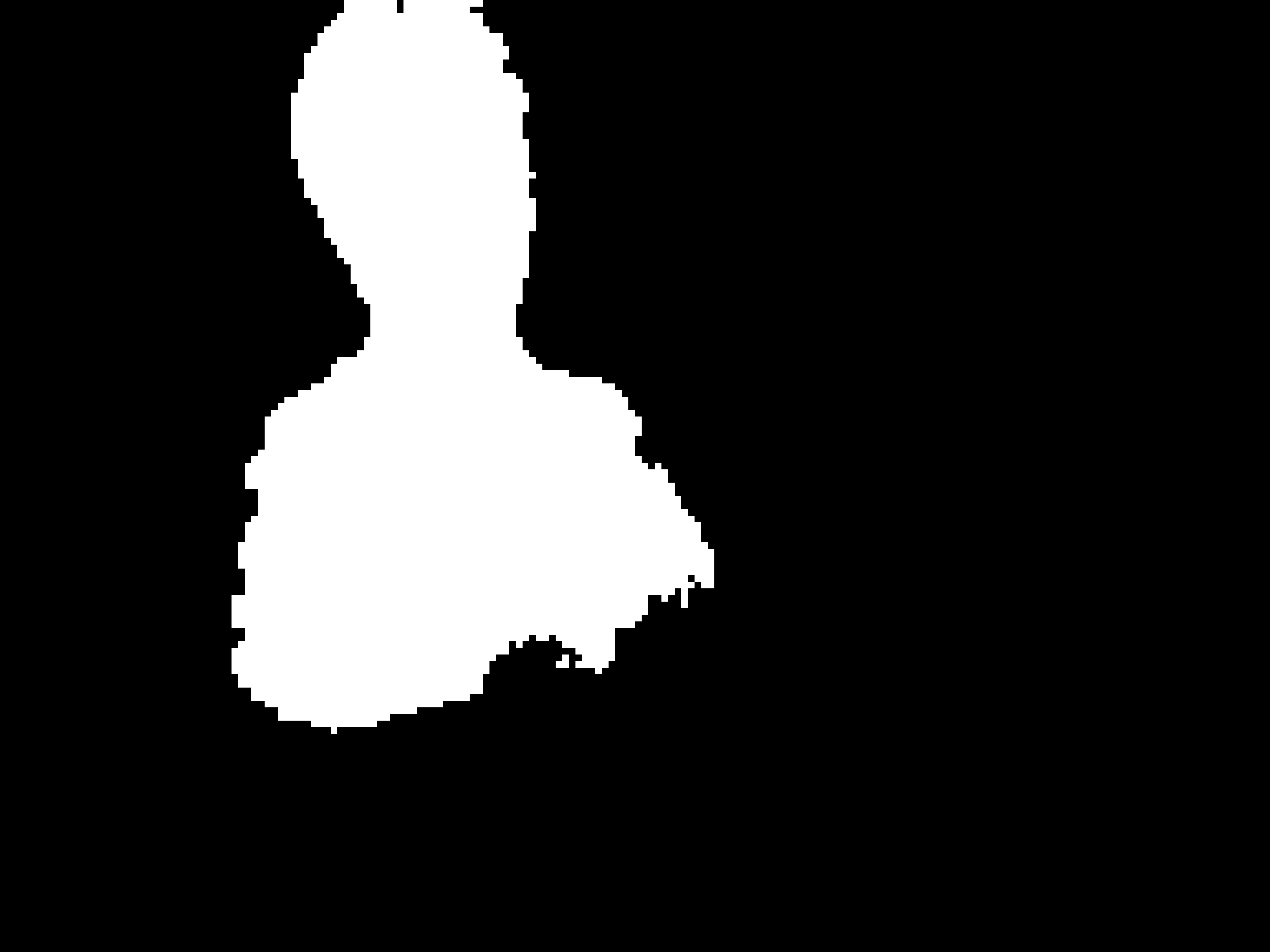}%
        \label{subfig:example_FP_BP_mask}}
        \caption{Example portrait images with overlaid BP-driven NCC maps (a, e), relative depth maps (b, f) and Otsu thresholding masks (c, g) from \cite{BARACCHI_2021}. Binary masks $\mathbf{M}^{\text{(PRNU)}}$ (d, h) are obtained after thresholding NCC maps in (a, e).}
        \label{fig:depth_baracchi}
\end{figure}
 
Observations from the portrait images used in this experiment suggest that Apple's algorithm incorporates more than just the depth map, likely leveraging additional information (such as the auxiliary mattes mentioned in Sect.~\ref{subsec:apple_portrait_mode}) to decide which region in the image is finally blurred. This is evident in the example (taken from the reference set) shown in Fig.~\ref{subfig:example_misdetection_user2}, where its depth map (Fig.~\ref{subfig:depth_map}) does not capture details of the girl's eyes, yet Apple's algorithm keeps only the eye region unblurred. The binary mask produced by Baracchi \emph{et al.}'s method, obtained via Otsu thresholding (Fig.~\ref{subfig:baracchi_mask}), fails to exclude many SDNP-containing regions during PRNU extraction, leading to performance degradation. In contrast, our approach produces a binary mask $\mathbf{M}^{\text{(PRNU)}}$ (Fig.~\ref{subfig:SDNP_mask}) that effectively minimizes the BP leakage into the PRNU estimate $\hat{\mathbf{K}}$. The NCC values computed between the residues and the BP after applying the respective masks are $0.3156$ for the weighted approach and $0.0201$ for ours.

Another example of depth map-related issues comes from the negative class of the test set (Fig.~\ref{subfig:example_false_positive}) and its corresponding depth map (Fig.~\ref{subfig:example_FP_depth_map}). Otsu thresholding produces a binary mask (Fig.~\ref{subfig:example_FP_binary_mask}) that fails to segment the in-focus region, instead isolating an area composed almost entirely of the SDNP. This results in a false positive, with a test statistic of $\eta=1.71\times10^6$. In contrast, our BP-driven mask (Fig.~\ref{subfig:example_FP_BP_mask}) yields $\eta^\prime=4.62$, which aligns with the expected value for a negative sample. These findings highlight that minimizing BP leakage during both PRNU extraction and subsequent detection steps is crucial for improving source camera verification performance when dealing with Apple portrait images.

\subsubsection{Impact of BP Quality on Detection Performance}
\label{subsubsec:BP_quality_detection_performance}

We evaluate the impact of BP estimation quality using three estimates from our iPhone 15 (D81): $\hat{\mathbf{P}}_{\text{NL}}$ (Hi) with $L=110$, $\hat{\mathbf{P}}_{\text{SLM}}$ (Lo) with $L=10$, and $\hat{\mathbf{P}}_{\text{NL}}$ (Misc) with $L=158$, following the capture conditions in \cite[Sect.~4]{SUP_MATERIAL_2025}. The PRNU is estimated using 10 reference portrait images from the same device. The test set consists of 100 portrait images from the iPhone 15 (positive class) and 100 from an iPhone 13 with iOS~16 (negative class), sourced from Flickr. Both devices share the same BP \mytextcircled{6} (Tab.~\ref{tab:diff_BP_versions}).

In this experiment, we ensured that the images used for PRNU extraction produced a PRNU estimate $\hat{\mathbf{K}}_{\mathbf{b}}$ with non-null support, making it suitable for use with the binary method from \cite{BARACCHI_2021}. Under these conditions, the binary method outperforms both the weighted and baseline approaches in AUC, whereas our approach achieves the best performance, with minimal variation across the three BP estimates. Among them, $\hat{\mathbf{P}}_{\text{NL}}$ (Hi) achieves the highest AUC, followed by $\hat{\mathbf{P}}_{\text{NL}}$ (Misc) and $\hat{\mathbf{P}}_{\text{SLM}}$ (Lo). Notably, $\hat{\mathbf{P}}_{\text{NL}}$ (Misc) is preferable in low False Positive Rate (FPR) scenarios, maintaining the highest partial AUC for $\text{FPR}\leq0.05$ ($0.0373$), closely followed by $\hat{\mathbf{P}}_{\text{NL}}$ (Hi) ($0.0372$). While $\hat{\mathbf{P}}_{\text{SLM}}$ (Lo) consistently underperforms, its ease of extraction makes it a practical alternative.

Based on \cite[Fig.~8]{SUP_MATERIAL_2025} and \cite[Sect.~4]{SUP_MATERIAL_2025}, larger differences in detection performance across BP estimates might be expected. However, while those results were obtained using uniform image blocks in controlled conditions, this experiment processes real-world scenes. Here, applying a fixed threshold of $\alpha=0.07$ across all BP estimates produced highly similar binary masks $\mathbf{M}^{\text{(PRNU)}}$, resulting in nearly identical PRNU estimates $\mathbf{K}^\prime$ and comparable residue masking in \eqref{eq:eta_prime}, ultimately yielding similar detection performance. Future work could explore adaptive thresholds tailored to each BP estimate to better capture variations in BP quality.

\subsubsection{Comparative Analysis with iPhone 12 mini and Multiple Users}
\label{subsubsec:comp_analysis_iphone12mini_multiuser}

In this experiment, we use the BP estimate $\hat{\mathbf{P}}_{\text{NL}}$ ($L=50$) from our iPhone 12 mini (D51) and extract the PRNU from 50 reference portrait images of the same device. The positive class consists of 60 portrait images from the same iPhone 12 mini, while the negative class includes 60 images from Flickr: 37 and 5 from two iPhone 12, and 10 and 8 from two other iPhone 12 mini. These models share the same BP~\mytextcircled{5} (Tab.~\ref{tab:diff_BP_versions}).

The results in Fig.~\ref{subfig:ROC_iphone12mini} follow the same trend as those in Fig.~\ref{subfig:ROC_iphone15} for the iPhone 15. However, unlike the first experiment with iPhone 11 Pro images from \cite{IULIANI_2021} (Fig.~\ref{subfig:ROC_iphone11pro}), perfect detection is not achieved by our approach for newer devices (iPhone 12 mini and iPhone 15). A key factor is the iOS version: the iPhone 11 Pro images were captured with iOS 13, whereas the newer devices use iOS 17. This suggests that Apple has refined SDNP embedding over time, making it more selective (particularly by omitting certain areas like edges). This is evident in the NCC maps in Fig.~\ref{fig:BP_corr_map}, where the iPhone 11 Pro Max image (iOS 13, Fig.~\ref{subfig:01_albisani_example}) shows less correlation loss at the edges compared to the iPhone 15 image (iOS 17, Fig.~\ref{subfig:02_albisani_example}). In the latter case, correlation loss is pronounced in the brightness transitions on the ColorChecker, affecting scene segmentation for PRNU matching. These areas, still blurred, likely retain fewer PRNU traces, reducing detection accuracy.

\subsection{Multi-Device BP-aware PRNU Matching}
\label{subsec:multi_device}

To assess the performance of the proposed BP-aware PRNU matching approach across multiple devices sharing identical or distinct BPs, we extended the experiments to eight smartphones: two devices from each of four models (iPhone 7 Plus, X, 11 Pro, and 12 mini). The selected devices, as listed in \cite[Tab.~6]{SUP_MATERIAL_2025}, are: two iPhone 7 Plus (D02 and D06) sharing BP~\mytextcircled{1} (note that D02 also includes images with BP~\mytextcircled{2} under iOS~11); two iPhone X (D15 and D16) sharing BP~\mytextcircled{4}; two iPhone 11 Pro (D36 and D37), also with BP~\mytextcircled{4}; and two iPhone 12 mini (D48 and D51) sharing BP~\mytextcircled{5}. For each device, the PRNU was extracted from a subset of portrait images: 50 for D02 (out of 311), 50 for D06 (out of 118), 15 for D15 (out of 43), 50 for D16 (out of 140), 15 for D36 (out of 62), 15 for D37 (out of 28), 15 for D48 (out of 27), and 40 for D51 (out of 110). The remaining images were used for testing. For cross-device tests, all available portrait images were used as test data.

\begin{figure}[t]
    \centering
    \includegraphics[width=\linewidth]{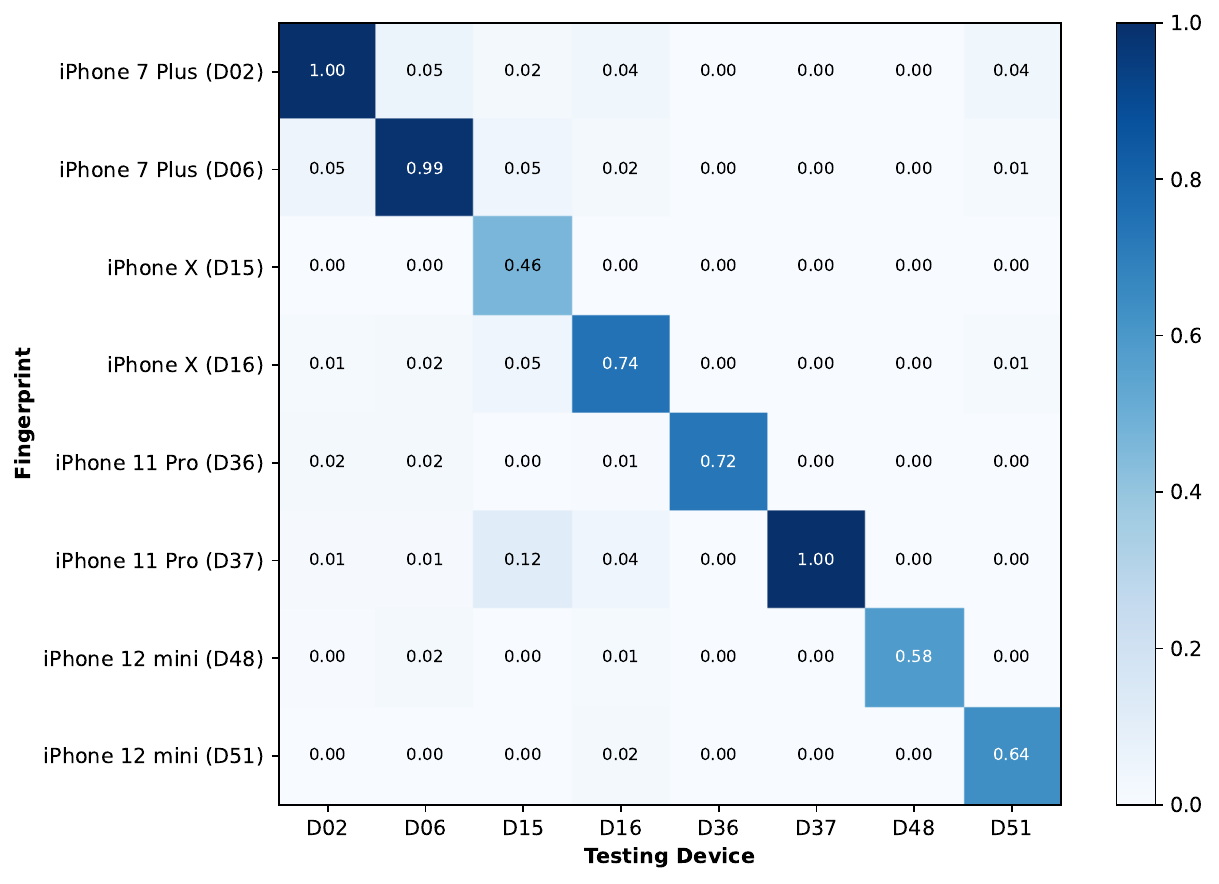}
    \caption{TPR-FPR performance matrix for BP-aware PRNU matching across eight devices (two per model: iPhone 7 Plus, X, 11 Pro, and 12 mini). Each entry corresponds to an independent binary test: diagonal elements report TPRs and off-diagonal elements report FPRs, computed at a fixed threshold $\tau^\prime=160$.}
    \label{fig:cmatrix}
\end{figure}

Under this setup, we applied our BP-aware PRNU matching method and computed $\eta^\prime$ scores for each PRNU fingerprint/device pair. The resulting confusion matrix, showing TPRs on the diagonal and FPRs elsewhere, is presented in Fig.~\ref{fig:cmatrix}. Results were obtained with a fixed threshold $\tau^\prime=160$, selected as a good trade-off based on Sect.~\ref{subsubsec:comp_analysis_iuliani}. Using the conventional PRNU verification threshold $\tau^\prime=60$ would increase the average TPR from $0.77$ to $0.91$ but also raise the average FPR from $0.011$ to $0.05$.

Overall, the proposed BP-aware approach keeps mismatched images under control at an overall operational point of $\text{FPR}=0.011$, though some low TPRs remain (e.g., $0.46$ for D15 and $0.58$ for D36), indicating room for improvement. A particularly high FPR (0.12) is observed between iPhone X (D15) and iPhone 11 Pro (D37), suggesting that part of the shared BP~\mytextcircled{4} leaks into the extracted PRNU. Additionally, low TPRs are partly due to portrait images with small non-bokeh regions with respect to the 12MP frame, where the estimated similarity is less reliable.

\subsection{Robustness of BP Detection under Post-Processing}
\label{subsec:robustness_postproc}

\begin{figure}[t]
    \centering
    \includegraphics[height=4.8cm]{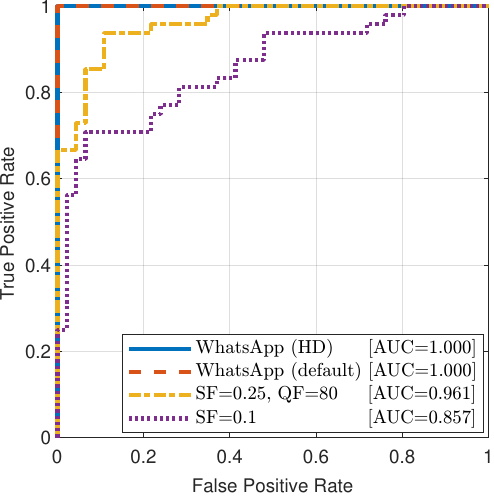}
    \caption{ROC curves showing BP detection robustness under post-processing. WhatsApp-shared images (default and HD modes) achieve perfect detection, while performance degrades under strong and extreme downscaling.}
    \label{fig:ROC_whatsapp}
\end{figure}

Finally, we evaluated the robustness of our BP detection approach against post-processing by testing Apple portrait images shared through WhatsApp. We sent 48 portrait images from our iPhone 12 mini using two modes: the default mode, which reduces resolution (12MP to 2MP) and applies compression, and the HD mode, which preserves resolution with moderate compression. These formed the positive classes, while the negative classes included 46 non-portrait images shared under the same conditions. As shown in Fig.~\ref{fig:ROC_whatsapp}, both modes had no impact on BP detection, achieving perfect results comparable to cases without post-processing. To illustrate a more challenging case where detection performance starts to drop, we also included ROC curves for images downscaled to a Scaling Factor (SF) of $0.25$ and JPEG compressed with a Quality Factor (QF) of 80, as well as for extreme image downscaling by $\text{SF}=0.1$ without additional compression. Further details on the effects of scaling and compression can be found in \cite[Sect.~5.6]{SUP_MATERIAL_2025}.

\section{Conclusions}
\label{sec:conclusions}

The distinctive Apple's SDNP and its underlying BP have been comprehensively analyzed in this paper, characterizing their role in iPhone portrait images and their forensic implications. We first proposed BP extraction methods under different lighting conditions and then examined their dependencies on luminance, ISO settings, and software variations. Our findings show that SDNP knowledge enhances forensic tasks, including portrait-mode image traceability and improved PRNU-based source verification in realistic forensic scenarios.

Future work will extend this analysis to additional Apple devices not covered in this study (e.g., iPads and other iPhone models running different iOS versions), as well as Apple-related software such as \textit{Aperture} (now discontinued) and \textit{Photos}. Since BP detection, identification, and localization performance depend on the quality of the reference BPs, we will continue refining these references to fully exploit the potential of the proposed approach. In addition, future research will explore whether the characteristic arc patterns observed in BP-driven NCC maps, both between iPhone BPs (e.g., \mytextcircled{5} × \mytextcircled{6}) and between iPhone and Apple software–related BPs, such as \textit{Photos} (e.g., \mytextcircled{6} × \mytextcircled{8}), hold forensic significance. These patterns may offer valuable insights for distinguishing specific \textit{Photos} software or iOS versions within the same iPhone model (see, e.g., the iPhone 13 Series in Tab.~1), tracing algorithmic updates in Apple's image processing pipeline, and potentially even reverse-engineering the underlying mechanisms used to generate the different BPs. In parallel, although initial results show a marginal gain, we plan to further investigate the potential of the estimated brightness-dependent function $\hat{g}(\cdot)$ to uncover possible benefits or new applications. Additionally, we aim to investigate adaptive thresholding for SDNP localization and continue improving BP extraction and detection to optimize performance across varying image conditions and BP quality levels. Overall, we plan to advance forensic applications such as forgery localization using BP-aware PRNU analysis. Although the proposed proof-of-concept demonstrates potential, further refinement is needed to overcome current limitations, notably the low resolution of the PRNU-driven NCC map.

Finally, while this work focuses on Apple devices, an important open question is whether similar methods can generalize to other smartphone brands. In this regard, we recently reported distinct diagonal artifacts in images from certain Samsung devices \cite{VAZQUEZPADIN_2025}, which, while differing from Apple's BP (being present in standard photo mode but disappearing in portrait images where bokeh is applied) nonetheless support the concept of modeling device-specific synthetic patterns. Moreover, preliminary analyses suggest that some Huawei devices embed analogous patterns to Apple's BP within bokeh regions. These findings highlight the broader applicability of device-specific synthetic artifact modeling for forensic analysis. Or course, we cannot guarantee that such artifacts will persist in future versions of Apple's or other manufacturers' imaging pipelines.

\section*{Acknowledgments}
We would like to acknowledge all individuals who contributed portrait-mode images for this study.

\appendices

\section{Derivation of the LS Estimate of \texorpdfstring{$\hat{g}(\cdot)$}{g(·)}}
\label{app:MMSE_g_y}

In this appendix we derive the estimate of the brightness-dependent scaling function $\hat{g}(\cdot)$ following an LS approach. Let $T$ and $\text{tr}$ denote respectively the transpose and trace matrix operators. Given two matrices $\mathbf{A}, \mathbf{B}$ and a scalar $c$, let $\mathbf{E} \triangleq \left(\mathbf{A}-c\mathbf{B}\right)^T \left(\mathbf{A}-c\mathbf{B}\right)$. Notice that $\|\mathbf{A}-c\mathbf{B}\|_{\text{F}}^2=\text{tr}(\mathbf{E})$. Taking the derivative of $\mathbf{E}$ w.r.t. $c$, we have $\frac{d\mathbf{E}}{dc}=2c \mathbf{B}^T \mathbf{B} -\mathbf{B}^T\mathbf{A} - \mathbf{A}^T\mathbf{B}$. Noticing that $\text{tr}(\mathbf{A}^T \mathbf{B})=\langle \mathbf{A}, \mathbf{B}\rangle_{\text{F}}$ and that the trace operator commutes with the derivative, we have that $\frac{d \ \text{tr}(\mathbf{E})}{dc} = 2 c \|\mathbf{B}\|_{\text{F}}^2 - 2 \langle \mathbf{A},\mathbf{B}\rangle_{\text{F}}$.

Assuming that block $\mathbf{Z}_{m,t}$ contains the base pattern $\mathbf{P}_m$ everywhere, i.e. $\mathbf{M}_\text{(blur)}=\mathbf{1}$ and $\mathbf{M}'_\text{(blur)}=\mathbf{0}$, and that it has a flat background,  Eq.~\eqref{eq:Z} becomes  $\mathbf{Z}_{m,t}=\mathbf{Y}_{m,t}^\prime+\gamma_{\text{ISO}}\cdot g(y'_{m,t}) \mathbf{P}_m+\mathbf{\Phi}_{m,t}$. Here we make no assumptions on the mean and std of $\mathbf{P}$, as some estimates of $\mathbf{P}$ (e.g., $\hat{\mathbf{P}}_\text{SLM}$) may not share its second order statistics. We are interested in finding $y'_{m,t} \triangleq \mu(\mathbf{Y}^\prime_{m,t})$ and $g(y'_{m,t})$ minimizing
\begin{equation}
    \|\mathbf{Z}_{m,t}-\mathbf{Y}_{m,t}^\prime-\gamma_{\text{ISO}}\cdot g(y'_{m,t}) \mathbf{P}_m\|^2_{\text{F}}.
    \label{eq:Frobnorm}
\end{equation}
First, notice that this norm is minimized when $\mu\left(\mathbf{Z}_{m,t}-\mathbf{Y}_{m,t}^\prime-\gamma_{\text{ISO}}\cdot g(y'_{m,t}) \mathbf{P}_m\right)=0$, leading to
\begin{equation}
\label{eq:yhat}
\hat{y}'_{m,t}=z_{m,t}-\gamma_\text{ISO} \cdot g(y'_{m,t}) \cdot \mu(\mathbf{P}_m),
\end{equation}
where $z_{m,t} \triangleq \mu(\mathbf{Z}_{m,t})$. 
Now we use the algebraic result at the beginning of this Appendix to compute the derivative of \eqref{eq:Frobnorm} with respect to $g(y'_{m,t})$. Equating to zero, we obtain  
\begin{equation}
\label{eq:settozero}
\langle\mathbf{Z}_{m,t},\mathbf{P}_m \rangle_{\text{F}} - \langle \mathbf{Y}^\prime_{m,t}, \mathbf{P}_m \rangle_{\text{F}}-\gamma_\text{ISO} \cdot g(y'_{m,t}) \cdot \|\mathbf{P}_m\|_{\text{F}}^2=0.
\end{equation}
Since the block is assumed to have flat background, we can approximate $\langle \mathbf{Y}^\prime_{m,t}, \mathbf{P}_m \rangle_{\text{F}} \approx y'_{m,t} \cdot \mu(\mathbf{P}_m) B^2 \approx z_{m,t} \cdot \mu(\mathbf{P}_m) B^2-\gamma_\text{ISO} \cdot g(y'_{m,t}) \cdot \left(\mu(\mathbf{P}_m)\right)^2 B^2$, where the second approximation follows from substituting $y'_{m,t}$ in the first by its estimate in \eqref{eq:yhat}. Replacing $\langle \mathbf{Y}^\prime_{m,t}, \mathbf{P}_m \rangle_{\text{F}}$ in \eqref{eq:settozero} by this approximation, substituting  $\mathbf{P}_m$ and $\gamma_{\text{ISO}}$ by their respective estimates, and solving for $g(y'_{m,t})$, we obtain \eqref{eq:hat_g}. 

\bibliographystyle{IEEEtran}
\bibliography{SDNP_draft}







\vspace{-33pt}
\begin{IEEEbiography}[{\includegraphics[width=1in,height=1.25in,clip,keepaspectratio]{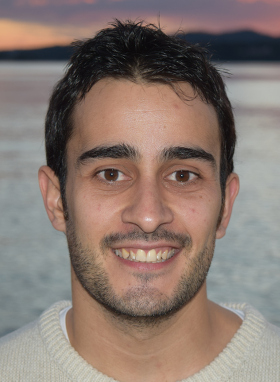}}]{David V\'azquez-Pad\'in} (Member, IEEE) received the M.S. degree in telecommunication engineering and the Ph.D. degree from the University of Vigo, Vigo, Spain, in 2008 and 2016, respectively.

He has been affiliated with the Signal Theory and Communications Department at the University of Vigo since 2008, where he currently holds a postdoctoral researcher position. In 2012, he was a visiting researcher at the Department of Information Engineering of the University of Florence, Italy, for a six-month period.

His research interests include multimedia forensics and digital watermarking, in which he has authored several publications in leading international journals and conferences and has one pending international patent application. He has actively contributed to the European research projects REWIND, NIFTY, and UNCOVER, and he is currently a member of the EURASIP BForSec Technical Area Committee.
\end{IEEEbiography}

\vspace{-33pt}
\begin{IEEEbiography}[{\includegraphics[width=1in,height=1.25in,clip,keepaspectratio]{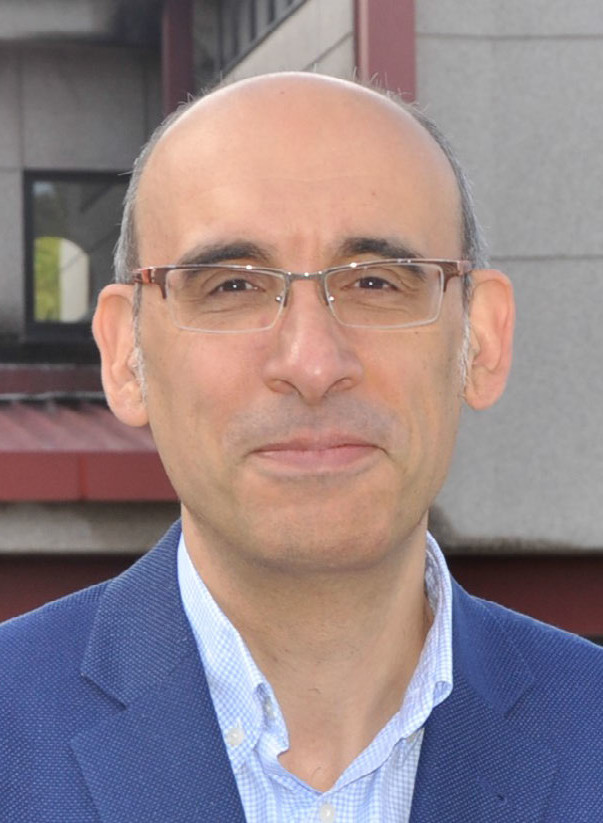}}]{Fernando P\'erez-Gonz\'alez} (Fellow, IEEE) received the degree in telecommunication engineering from the University of Santiago, Santiago, Spain, in 1990, and the Ph.D. degree in telecommunications engineering from the University of Vigo, Vigo, Spain, in 1993.

He is currently a Professor with the School of Telecommunication Engineering, University of Vigo. From 2007 to 2010, he was the Program Manager of the Spanish National Research and Development Plan on Electronic and Communication Technologies, Ministry of Science and Innovation. From 2009 to 2011, he was the Prince of the Asturias Endowed Chair of Information Science and Technology, The University of New Mexico, Albuquerque, NM, USA. From 2007 to 2014, he was the Executive Director of the Galician Research and Development Center in Advanced Telecommunications. He has been a Principal Investigator with the Signal Processing in Communications Group, University of Vigo, which participated in several European projects. He has coauthored over 70 papers in leading international journals, 180 peer-reviewed conference papers, and several international patents. His research interests include the areas of digital communications, adaptive algorithms, privacy enhancing technologies, and information forensics and security.

Dr. P\'erez-Gonz\'alez has served as Associate Editor of IEEE Signal Processing Letters (2005–2009), IEEE Transactions on Information Forensics and Security (IEEE-TIFS, 2006–2010, 2023–present), and was Editor-in-Chief of the EURASIP International Journal on Information Security (2017–2022). From 2019 to 2021, he was Senior Area Editor for IEEE-TIFS. 
\end{IEEEbiography}

\vspace{-33pt}
\begin{IEEEbiography}[{\includegraphics[width=1in,height=1.25in,clip,keepaspectratio]{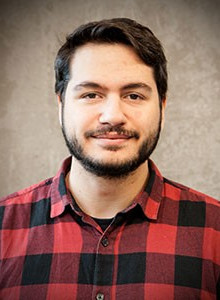}}]{Pablo P\'erez-Migu\'elez} received the B.S. and M.S. degrees in telecommunication engineering from the University of Vigo, Vigo, Spain, in 2017 and 2019, respectively.

He has been working at the Signal Theory and Communications Department in the University of Vigo since 2019 as a research engineer, mainly focused on technology transfer projects.

His research interests include multimedia forensics and synthetic media detection.
\end{IEEEbiography}

\end{document}